\pdfoutput=1

\documentclass[11pt]{article}

\usepackage[final]{acl}

\usepackage{times}
\usepackage{latexsym}
\usepackage{lipsum}

\usepackage[T1]{fontenc}

\usepackage[utf8]{inputenc}

\usepackage{microtype}

\usepackage{inconsolata}

\usepackage{graphicx}

\usepackage{amssymb}
\usepackage{tabularray}
\usepackage{amsmath}
\usepackage{booktabs}
\usepackage{multirow}
\usepackage{graphicx}
\usepackage{array}
\usepackage{ragged2e}
\usepackage{colortbl}
\usepackage{subcaption}
\usepackage[normalem]{ulem}
\usepackage{CJKutf8}
\usepackage{algorithm}
\usepackage{algpseudocode}
\usepackage{tcolorbox}
\usepackage{pgf} 

\definecolor{customred}{rgb}{0.85, 0.25, 0.25}
\definecolor{customorange}{rgb}{0.9, 0.5, 0.2}
\definecolor{customyellow}{rgb}{0.9, 0.8, 0.2}
\definecolor{customgreen}{rgb}{0.5, 0.85, 0.5}
\definecolor{customgreenblue}{rgb}{0.36, 0.58, 0.66}
\definecolor{customblue}{rgb}{0.3, 0.4, 0.85}
\newcommand{\redcell}[1]{%
    \pgfmathsetmacro\percent{125*((#1) - 0.8)} 
    \edef\x{\noexpand\cellcolor{customred!\percent!white}}\x#1
}

\newcommand{\orangecell}[1]{%
    \pgfmathsetmacro\percent{125*((#1) - 0.63)} 
    \edef\x{\noexpand\cellcolor{customorange!\percent!white}}\x#1
}

\newcommand{\yellowcell}[1]{%
    \pgfmathsetmacro\percent{125*((#1) - 0.15)} 
    \edef\x{\noexpand\cellcolor{customyellow!\percent!white}}\x#1
}

\newcommand{\greencell}[1]{%
    \pgfmathsetmacro\percent{125*((#1) - 0.63)} 
    \edef\x{\noexpand\cellcolor{customgreen!\percent!white}}\x#1
}

\newcommand{\greenbluecell}[1]{%
    \pgfmathsetmacro\percent{125*((#1) - 0.16)} 
    \edef\x{\noexpand\cellcolor{customgreenblue!\percent!white}}\x#1
}

\newcommand{\bluecell}[1]{%
    \pgfmathsetmacro\percent{125*((#1) - 0.44)} 
    \edef\x{\noexpand\cellcolor{customblue!\percent!white}}\x#1
}

\newcommand{\orangecelldata}[1]{%
    \pgfmathsetmacro\percent{125*((#1) - 0.64)} 
    \edef\x{\noexpand\cellcolor{customorange!\percent!white}}\x#1
}

\newcommand{\yellowcelldata}[1]{%
    \pgfmathsetmacro\percent{125*((#1) - 0.2)} 
    \edef\x{\noexpand\cellcolor{customyellow!\percent!white}}\x#1
}

\newcommand{\greencelldata}[1]{%
    \pgfmathsetmacro\percent{125*((#1) - 0.6)} 
    \edef\x{\noexpand\cellcolor{customgreen!\percent!white}}\x#1
}

\newcommand{\greenbluecelldata}[1]{%
    \pgfmathsetmacro\percent{125*((#1) - 0.15)} 
    \edef\x{\noexpand\cellcolor{customgreenblue!\percent!white}}\x#1
}

\newcommand{\redcellfull}[1]{%
    \pgfmathsetmacro\percent{125*((#1) - 0.65)} 
    \edef\x{\noexpand\cellcolor{customred!\percent!white}}\x#1
}

\newcommand{\orangecellfull}[1]{%
    \pgfmathsetmacro\percent{125*((#1) - 0.47)} 
    \edef\x{\noexpand\cellcolor{customorange!\percent!white}}\x#1
}

\newcommand{\yellowcellfull}[1]{%
    \pgfmathsetmacro\percent{125*((#1) - 0.09)} 
    \edef\x{\noexpand\cellcolor{customyellow!\percent!white}}\x#1
}

\newcommand{\greencellfull}[1]{%
    \pgfmathsetmacro\percent{125*((#1) - 0.45)} 
    \edef\x{\noexpand\cellcolor{customgreen!\percent!white}}\x#1
}

\newcommand{\greenbluecellfull}[1]{%
    \pgfmathsetmacro\percent{125*((#1) - 0.1)} 
    \edef\x{\noexpand\cellcolor{customgreenblue!\percent!white}}\x#1
}

\newcommand{\bluecellfull}[1]{%
    \pgfmathsetmacro\percent{125*((#1) - 0.28)} 
    \edef\x{\noexpand\cellcolor{customblue!\percent!white}}\x#1
}

\title{\textsc{Cheems}: A Practical Guidance for Building and Evaluating \\ Chinese Reward Models from Scratch}

\author{
  \textbf{Xueru Wen\textsuperscript{1,2†}},
  \textbf{Jie Lou\textsuperscript{3†*}},
  \textbf{Zichao Li\textsuperscript{1,2†}},
  \textbf{Yaojie Lu\textsuperscript{1*}},
  \textbf{XingYu\textsuperscript{3}},
  \textbf{Yuqiu Ji\textsuperscript{3}},
  \textbf{Guohai Xu\textsuperscript{3}},
\\
  \textbf{Hongyu Lin\textsuperscript{1}},
  \textbf{Ben He\textsuperscript{1,2}},
  \textbf{Xianpei Han\textsuperscript{1}},
  \textbf{Le Sun\textsuperscript{1}},
  \textbf{Debing Zhang\textsuperscript{3}}
\\
  \textsuperscript{1}Chinese Information Processing Laboratory,
  Institute of Software, \\ Chinese Academy of Sciences, Beijing, China
\\
  \textsuperscript{2}University of Chinese Academy of Sciences, Beijing, China
\\
  \textsuperscript{3}Xiaohongshu Inc
\\
  \texttt{\{wenxueru2022,lizichao2022\}@iscas.ac.cn}
\\
  \texttt{\{luyaojie,hongyu,sunle,xianpei\}@iscas.ac.cn}
\\ 
  \texttt{benhe@ucas.edu.cn loujie0822@gmail.com dengyang@xiaohongshu.com}
}

\begin{document}
\maketitle
\renewcommand{\thefootnote}{}
\footnotetext{\textsuperscript{†}These authors contributed equally to this work.}
\footnotetext{\textsuperscript{*}Corresponding authors.}
\renewcommand{\thefootnote}{\arabic{footnote}}

\begin{abstract}
Reward models (RMs) are crucial for aligning large language models (LLMs) with human preferences.
However, most RM research is centered on English and relies heavily on synthetic resources, which leads to limited and less reliable datasets and benchmarks for Chinese.
To address this gap, we introduce CheemsBench, a fully human-annotated RM evaluation benchmark within Chinese contexts, and CheemsPreference, a large-scale and diverse preference dataset annotated through human-machine collaboration to support Chinese RM training.
We systematically evaluate open-source discriminative and generative RMs on CheemsBench and observe significant limitations in their ability to capture human preferences in Chinese scenarios.
Additionally, based on CheemsPreference, we construct an RM that achieves state-of-the-art performance on CheemsBench, demonstrating the necessity of human supervision in RM training.
Our findings reveal that scaled AI-generated data struggles to fully capture human preferences, emphasizing the importance of high-quality human supervision in RM development.
\end{abstract}

\begin{figure}[ht]
    \centering
    \includegraphics[width=\linewidth]{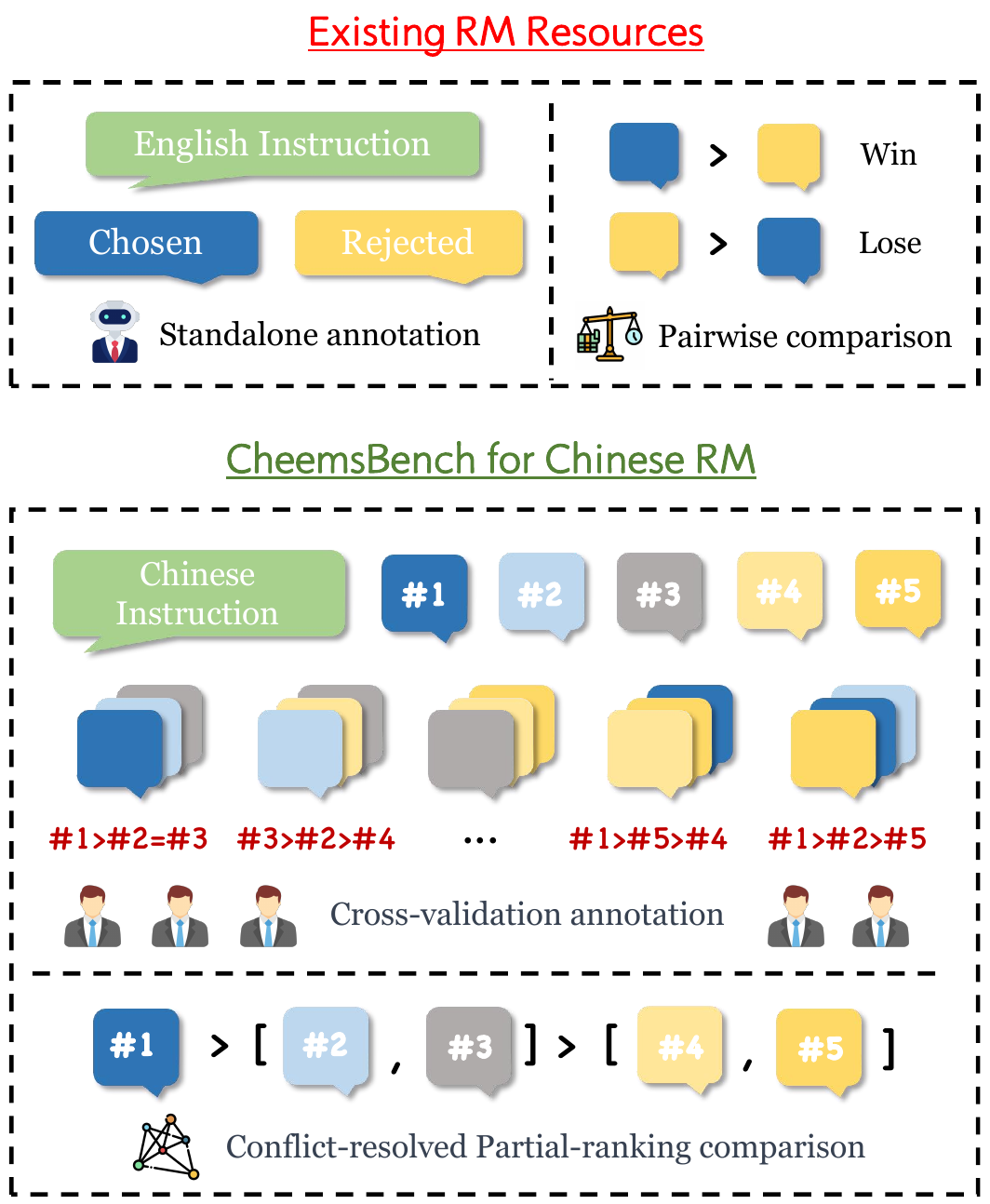}
    \caption{The differences in construction and usage between CheemsBench and the existing RM resources.}
    \label{fig:first_page}
\end{figure}

\section{Introduction}
With the rapid advancement of large language models \citep{qwen2,grattafiori2024llama3herdmodels}, post-training has emerged as a critical challenge to ensure their safety, reliability, and alignment with human values \citep{hou2024chatglmrlhfpracticesaligninglarge, lin2024baichuanalignmenttechnicalreport}. 
Reward models \citep{palan2019learningrewardfunctionsintegrating,ouyang2022traininglanguagemodelsfollow}, as core components of LLM post-training, play a pivotal role in capturing human preferences and guiding models to adhere more closely to human needs \citep{bai2022traininghelpfulharmlessassistant}.
By providing reward signals, RMs can guide parameter optimization during training \citep{ibarz2018rewardlearninghumanpreferences,ouyang2022traininglanguagemodelsfollow} or directly intervene in outputs during decoding\citep{khanov2024argsalignmentrewardguidedsearch, li2024cascaderewardsamplingefficient}.

\begin{table*}[t]
\centering
\resizebox{0.8\linewidth}{!}{%
\begin{tabular}{lccccc} 
\toprule
\multicolumn{1}{c}{\multirow{2}{*}{\textbf{Statistics}}} & \multicolumn{2}{c}{\textbf{CheemsBench}} &  & \multicolumn{2}{c}{\textbf{CheemsPreference}} \\ 
\cline{2-3}\cline{5-6}
\multicolumn{1}{c}{} & \textbf{Open Prompt} & \textbf{Human Instruction} &  & \textbf{GPT} & \textbf{Human} \\ 
\hline
\# Prompts & 1,146 & 1,346 &  & 27,861 & 3,260 \\
\# Responses & 5 & 5 &  & 5.29~ & 5.07~ \\
\# Comparisons & 7,838 & 9,762 &  & 332,370 & 37,618 \\
Avg. Char. of Prompt & 186.58~ & 197.04~ &  & 175.56~ & 164.08~ \\
Avg. Char. of Chosen & 437.50~ & 436.96~ &  & 457.92~ & 440.18~ \\
Avg. Char. of Rejected & 454.01~ & 446.43~ &  & 394.18~ & 432.84~ \\
\bottomrule
\end{tabular}
}
\caption{Statistics of CheemsBench and CheemsPreference: Number of prompts, average responses per prompt, comparisons (excluding ties), and average character lengths of prompts, chosen responses, and rejected responses.}
\label{tab:benchmark_stats}
\end{table*}

Despite the crucial role of RMs in post-training, current research is mainly focused on English.
For instance, Skywork-Reward \citep{liu2024skyworkrewardbagtricksreward} and UltraRM \citep{cui2023ultrafeedback} leverage high-quality English preference datasets \citep{zheng2023judging,ji2024pku} and benchmarks \citep{lambert2024rewardbenchevaluatingrewardmodels} to achieve superior performance. 
In contrast, the development of Chinese RMs faces significant challenges due to a lack of large-scale, high-quality preference datasets and comprehensive evaluation benchmarks.
Existing Chinese resources are often small in scale \citep{huozi,zhihu_rlhf_3k} and limited to specific domains \citep{Yang_Kyara_2024,DPO-zh-en-emoji2024,xu2023cvalues}, making them insufficient for LLM post-training.
Moreover, existing RM mainly rely on synthetic data, which struggles to accurately reflect human preferences.

To address this critical gap, this paper constructs a comprehensive and human-centric Chinese RM resource from scratch\footnote{\textsc{Cheems} stands for \textit{\b{C}\b{h}inese r\b{e}ward mod\b{e}l bench\b{m}ark and preference data\b{s}et}.}. 
It consists of two key datasets: 
(1) \textbf{CheemsBench}, a fully human-annotated and extensive Chinese RM evaluation benchmark that verifies whether RMs accurately capture and reflect human preferences; and (2) \textbf{CheemsPreference}, a large-scale, diverse Chinese preference dataset that provides supervised signals for training Chinese RMs, enabling them to effectively learn and model human preferences.

As shown in Figure \ref{fig:first_page}, unlike most RM resources that rely on machine-generated annotations \cite{zhou2024rmbcomprehensivelybenchmarkingreward}, CheemsBench and CheemsPreference are built on human supervision, thereby more accurately capturing realistic human values.
Moreover, while traditional RM benchmarks \cite{lambert2024rewardbenchevaluatingrewardmodels} typically rely on pairwise comparisons, recent studies \citep{wen2024rethinkingrewardmodelevaluation} have highlighted their limitations in reflecting downstream performances.
CheemsBench introduces a multi-response evaluation mechanism, which aligns closely with downstream tasks.

In CheemsBench, we combine open-source prompts and real-world human instructions with a comprehensive taxonomy to evaluate RM performance
To better align with downstream tasks and reduce preference-induced noise \citep{zhang2024divergingpreferencesannotatorsdisagree}, we sample five responses from various open- and closed-source LLMs for each prompt and conduct five rounds of human-driven triple-wise comparisons.
To address potential annotation conflicts, we design a graph-based conflict-resolving algorithm that generates unique and consistent partial rankings.
Using CheemsBench, we assess the progress of reward models and preference datasets in the Chinese context and identify considerable room for improvement in Chinese RMs. 

For CheemsPreference, we collect $27$k human instructions following a multi-tiered prompt taxonomy and sample more than $5$ responses per prompt from various LLMs, ensuring both prompt and response diversity. 
To alleviate inconsistencies and biases in GPT annotations \citep{stureborg2024largelanguagemodelsinconsistent} while reducing human effort, we design a distant supervision algorithm to improve data quality. 
Specifically, human annotators first label a small golden preference dataset, which is then used to train an RM to filter a larger GPT-annotated dataset.
The combined human- and GPT-annotated data form CheemsPreference, achieving state-of-the-art results on CheemsBench and performing well on the English RewardBench \cite{lambert2024rewardbenchevaluatingrewardmodels}.

Our contributions are summarized as follows:
\begin{itemize}
\item We propose CheemsBench, the first large-scale and comprehensive benchmark designed specifically for Chinese reward models.
\item We construct CheemsPreference, the first large-scale, diverse, and high-quality Chinese preference dataset.
\item We provide a comprehensive investigation into Chinese RM training and evaluation. The code and data associated with this work are available at \url{https://github.com/AlignRM/CheemsRM}.
\end{itemize}

\begin{figure*}[t]
    \centering
    \includegraphics[width=0.96\textwidth]{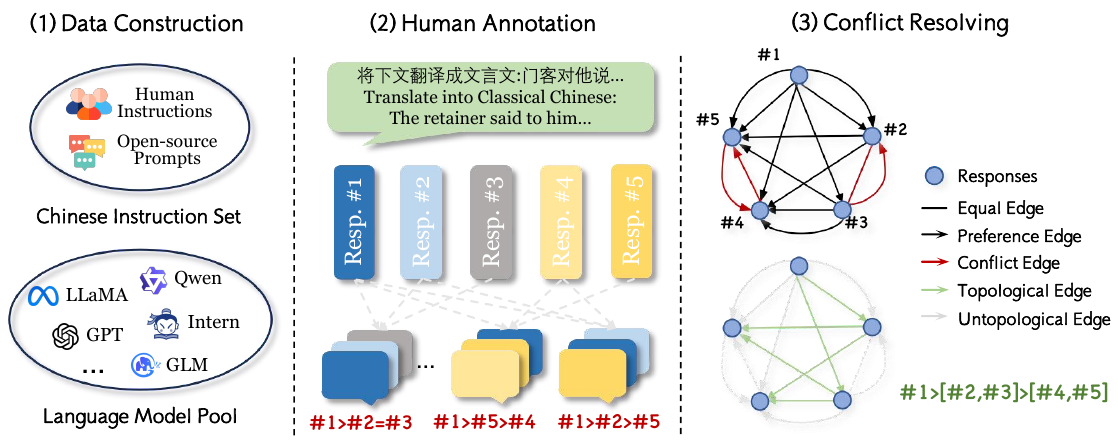}
    \caption{Chinese RM benchmark construction process. We utilize open-source prompts and human instructions and sample five responses from various models for each prompt. These responses then undergo five rounds of triple-wise manual comparisons. Unique partial rankings are generated by conflict resolving algorithm.}
    \label{fig:benchmark}
\end{figure*}

\section{Related Works}
\paragraph{Reinforcement Learning from Human Feedback.}
Reinforcement Learning from Human Feedback has been widely adopted for LLM alignment \citep{ouyang2022traininglanguagemodelsfollow,bai2022traininghelpfulharmlessassistant}.
Previous research mostly focuses on specific tasks like summarization \citep{stiennon2022learningsummarizehumanfeedback} and question answering \citep{nakano2022webgptbrowserassistedquestionansweringhuman}. 
Recent studies have expanded RLHF applications to broader domains \citep{hou2024chatglmrlhfpracticesaligninglarge,lin2024baichuanalignmenttechnicalreport,yu2024codepmpscalablepreferencemodel}, improving LLMs to be more helpful, honest, and harmless. 
RLHF enables models to align with human expectations more closely by integrating human preferences captured by reward models \citep{10.5555/645529.657801,brown2019deepbayesianrewardlearning,palan2019learningrewardfunctionsintegrating}. 
Thus, a reward model that accurately reflects human preferences is fundamental to the RLHF methodology.

\paragraph{Reward Model Training and Evaluation.}
To develop a RM that captures human preferences, current works gather preference data through manual annotation \citep{bai2022traininghelpfulharmlessassistant,zheng2023judging} or distilling advanced LLMs \citep{starling2023,cui2023ultrafeedback}.
These works mostly focus on English, overlooking Chinese contexts.
Existing Chinese preference datasets are generally small \citep{huozi,zhihu_rlhf_3k} or limited to specific tasks \citep{Yang_Kyara_2024,DPO-zh-en-emoji2024,xu2023cvalues}.
Beyond the training data, RM evaluation is also critical for post-training. 
The typical RM evaluation computes accuracy on a fixed test dataset \citep{lambert2024rewardbenchevaluatingrewardmodels}.
Recent studies \citep{son2024llmasajudgerewardmodel,kim2024evaluatingrobustnessrewardmodels,zhou2024rmbcomprehensivelybenchmarkingreward,liu2024rmbenchbenchmarkingrewardmodels,frick2024evaluaterewardmodelsrlhf,gureja2024mrewardbenchevaluatingrewardmodels} have attempted to strengthen the correlation with downstream performance. 
However, these benchmarks focus on English, raising questions about their applicability to Chinese contexts.

\section{Chinese RM Benchmark}
In this section, we introduce \textbf{CheemsBench}, a benchmark designed to comprehensively evaluate Chinese RMs. 
Our benchmark is characterized by: 
\textit{(1) High coverage}: We incorporate a wide range of prompts and sampling models, ensuring broad evaluation across diverse scenarios.
\textit{(2) High-quality annotation}: We derive a reliable preference ranking through multiple rounds of manual triple-wise comparisons and conflict resolving.
Figure \ref{fig:benchmark} illustrates the overall construction process.

\subsection{Data Construction}
\paragraph{Prompt Collection.}
We sample Chinese prompts from various open datasets, including Humaneval-XL \citep{peng2024humanevalxlmultilingualcodegeneration}, MathOctopus \citep{chen2024breakinglanguagebarriersmultilingual}, GAOKAO-Bench \citep{zhang2024evaluatingperformancelargelanguage}, HalluQA \citep{DBLP:journals/corr/abs-2310-03368}, Flames \citep{huang2023flames}, CLiB \citep{chinese-llm-benchmark}, AlignBench \citep{liu2023alignbench}, and COIG-CQIA \citep{COIG-CQIA}. 
We manually map their original categories into a unified system shown in Figure \ref{fig:open_label}.
We also include real-world human instructions for out-of-distribution evaluation.
To ensure thorough converge across different scenarios, we build a comprehensive categorization system as illustrated in Figure \ref{fig:close_label}. 
In total, we select 1,146 prompts from open-source datasets and 1,346 from human instructions.

\paragraph{Responses Collection.}
To ensure a wide range of response quality and distribution, we sample 5 responses per prompt from various models.
(1) Open-source models: Qwen2-7B/72B-Instruct \citep{qwen2}, Meta-Llama-3.1-8B/70B-Instruct \citep{grattafiori2024llama3herdmodels}, Llama3.1-8B/72B-Chinese-Chat \citep{shenzhi_wang_2024}, Internlm2-chat-1.8b \citep{cai2024internlm2technicalreport}, and GLM-4-9b-chat \citep{glm2024chatglm};
(2) Proprietary models: GPT-4 \citep{openai2024gpt4technicalreport}, GPT-3.5-turbo, GPT-4-turbo, and Claude-3-5-sonnet \citep{claude35sonnet}.
We observe that some open-source models demonstrate limited Chinese capabilities and tend to exhibit code-switching or even significant garbling\footnote{The LLaMA series shows a higher tendency for code-switching and nonsensical output, possibly due to its tokenizer vocabulary and insufficient training on Chinese corpora.}.
In such cases, we rely on human annotators to filter these responses during the annotation process. 
Specifically, annotators are instructed to discard responses containing substantial sections of meaningless content, while retaining those with minor code-switching that do not compromise semantic meaning.
This procedure allows us to account for LLMs' code-switching behavior during RM evaluation.

\subsection{Benchmark Labeling}
\label{sec:human_anno}
\paragraph{Human Annotation.} To accurately capture human preferences, CheemsBench relies entirely on human judgment for its annotation process.
Given a prompt and its corresponding 5 responses, we pre-design five annotation tasks, each comprising a triple-wise comparison of three adjacent responses. These tasks are distributed to different annotators who perform preference comparisons independently. All annotation results are then used to construct a ranked list of responses.

\paragraph{Conflict Resolving.} 
However, conflicts may arise due to the human preferences ambiguity and potential annotation errors. 
To derive reliable results, we develop a dedicated conflict resolving algorithm, as shown in Algorithm \ref{algo:resolve}. Specifically, we first transform the annotation results into a directed preference graph, where responses and preferences represent nodes and edges respectively. We then employ depth-first search to identify cycles in the graph, which indicate conflicts. These cycles are merged into larger nodes, and this process is repeated until no cycles remain in the graph. Finally, we perform topological sorting to obtain a partial ranking\footnote{Details about the algorithms and annotators are provided in Appendix \ref{sec:conflict_resolve} and Appendix \ref{sec:anno_details}, respectively.}.


\subsection{Evaluation Metrics}
Given multiple responses per prompt, there are many potential metrics for evaluation \citep{wen2024rethinkingrewardmodelevaluation}.
We first convert a partial ranking into multiple pair-wise comparisons and evaluate the accuracy as in the typical setting 
\citep{lambert2024rewardbenchevaluatingrewardmodels}:
\begin{equation}
    \text{Accuracy} = \frac{1}{N}\sum_{i=1}^{N}\mathbb{I}(r_w^i>r_l^i)
\end{equation}
where $N$ is the total number of pair-wise comparisons after transformation, and the indicator function $\mathbb{I}$ checks if the reward score for the preferred response $r_w^i$ is greater than that of its counterpart $r_l^i$.
Additionally, the exact match rate can be employed, which measures the proportion of prompts where all pair-wise comparisons are correctly sorted:
\begin{equation}
    \text{Exact Match} = \frac{1}{M}\sum_{j=1}^{M}\mathbb{I}\left(\bigwedge_{k} (r_w^{j,k} > r_l^{j,k})\right)
\end{equation}
where $M$ is the number of prompts, and the indicator function verifies if all comparisons are ordered correctly.
We obtain the final result by averaging the metrics from subsets of different categories.

\begin{figure*}[t]
    \centering
    \includegraphics[width=0.96\textwidth]{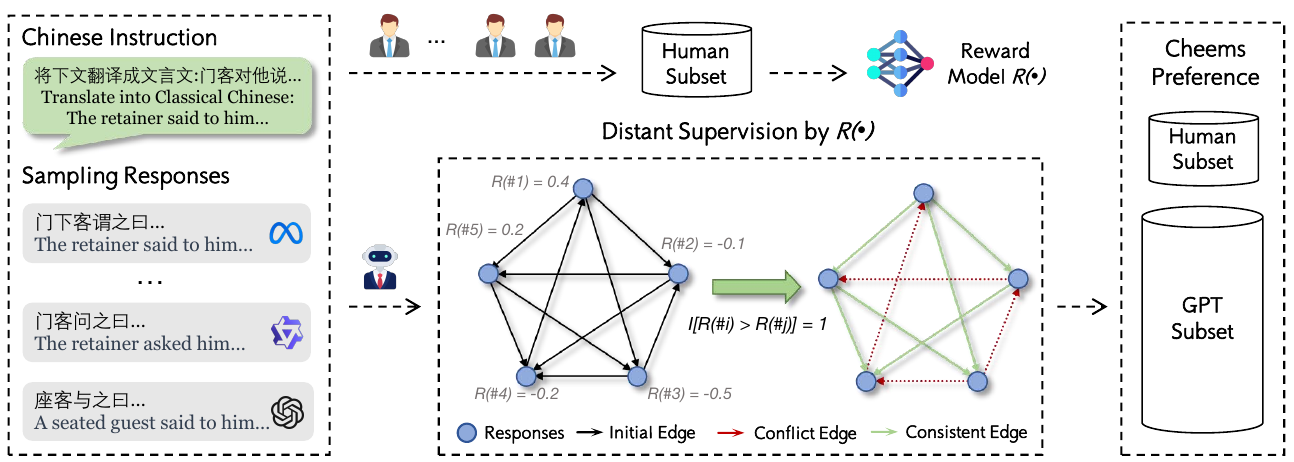}
    \caption{Chinese preference dataset construction process. Each prompt’s different responses and their annotation results form a directed graph. Circles in this preference graph indicate conflicts. We utilize the reward model trained on the human-annotated dataset to filter GPT annotations, thereby producing a directed acyclic graph.}
    \label{fig:preference_dataset}
\end{figure*}

\section{Chinese Preference Dataset}
In this section, we present the construction of \textbf{CheemsPreference}, as depicted in Figure \ref{fig:preference_dataset}. 
Our dataset is characterized by:
\textit{(1) Scale and diversity}: We amass $27$k real human instructions, featuring a comprehensive multi-tier categorization system, and sample multiple responses from a variety of models for each prompt.
\textit{(2) High-quality annotation}: 
We employ a distant supervision algorithm, which integrates both human annotations and GPT-4o to establish reliable partial preference ranks.

\subsection{Data Construction}

\paragraph{Prompt Collection.}
Diverse and high-quality instruction data are crucial for ensuring the robustness of RMs. 
To this end, we collect 27,861 real-world human instructions. 
To ensure extensive coverage of downstream scenarios, we develop a comprehensive multi-tier categorization system, which encompasses eight main categories with dozens of refined subcategories, as illustrated in Figure \ref{fig:preference_label}.

\paragraph{Response Collection.}
We sample responses from a broad range of models:
(1) Open-source models: Qwen2-7B/72B-Instruct \citep{qwen2}, Qwen2.5-7B/14B/32B/72B-Instruct \citep{qwen2.5}, Meta-Llama-3.1-8B/70B-Instruct \citep{grattafiori2024llama3herdmodels}, Llama3.1-8B/72B-Chinese-Chat \citep{shenzhi_wang_2024}, Internlm2-chat-1.8b \citep{cai2024internlm2technicalreport}, and GLM-4-9b-chat \citep{glm2024chatglm}.
(2) Proprietary models: GPT-4 \citep{openai2024gpt4technicalreport}, GPT-3.5-turbo, GPT-4-turbo, GPT-4o, and Claude-3-5-sonnet \citep{claude35sonnet}.
To guarantee the quality of responses, we implement rule-based methods to detect responses that are abnormally lengthy or contain excessive non-Chinese symbols.
Although this approach may have lower accuracy for prompts involving math or code, we prioritize a high recall rate to filter out more low-quality responses.
Finally, each prompt has more than $5$ responses on average.

\subsection{Distant Supervision}
\label{sec:preference_annotation}
The quality of preference data \citep{gao2024impactpreferencenoisealignment} is essential for the training of RM. 
While human annotation ensures high quality, it is expensive and challenging to obtain in large quantities. 
Conversely, GPT-based annotation is scalable but often inconsistent and biased \citep{stureborg2024largelanguagemodelsinconsistent}. 
To construct large-scale, high-quality Chinese preference data, we implement a distant supervision strategy for annotation.
We initially engage human annotators to label a small subset of data, following the protocol detailed in Section \ref{sec:human_anno}.
Subsequently, GPT-4o is employed to annotate a larger dataset.
For a set of $N$ responses, GPT-4o performs $C_{N}^{2}$ pair-wise comparisons between each response pairs\footnote{Annotation prompts can be found in Appendix \ref{sec:preference_prompt}.}.
To mitigate positional bias \citep{li2024splitmergealigningposition}, the order of responses in each comparison is randomized.
Although these GPT-4o annotations can exhibit inconsistencies, i.e., cycles in the preference graph, we employ an RM trained on human-annotated data to filter these annotations and establish a consistent partial order.
Additionally, we propose a length-debias post-hoc filtering strategy to alleviate length bias \citep{dubois2024lengthcontrolledalpacaevalsimpleway}.
This involves dividing the dataset into two groups, where the chosen response is longer or shorter than the rejected one, and downsampling the larger group to balance the dataset.

\renewcommand{\arraystretch}{0.75}
\begin{table*}[t]
\centering
\setlength{\extrarowheight}{0pt}
\addtolength{\extrarowheight}{\aboverulesep}
\addtolength{\extrarowheight}{\belowrulesep}
\setlength{\aboverulesep}{0pt}
\setlength{\belowrulesep}{0pt}
\resizebox{0.9\linewidth}{!}{%
\begin{tabular}{lcccccc} 
\toprule
\multicolumn{1}{c}{\multirow{2}{*}{\textbf{Model Name}}} & \multicolumn{1}{l}{\multirow{2}{*}{\textbf{RewardBench}}} & \multicolumn{2}{c}{\textbf{Open Prompt}} & \multicolumn{2}{c}{\textbf{Human Instruction}} & \multirow{2}{*}{\textbf{Overall}} \\ 
\cline{3-6}
\multicolumn{1}{c}{} & \multicolumn{1}{l}{} & Acc. & Exact. & Acc. & Exact. &  \\ 
\midrule
\multicolumn{7}{c}{\textbf{\textit{Generative Models as Reward Models}}} \\ 
\midrule
\href{https://huggingface.co/Skywork/Skywork-Critic-Llama-3.1-70B}{Skywork-Critic-Llama-3.1-70B} & \redcell{0.933} & \orangecell{0.755} & \yellowcell{0.320} & \greencell{0.731} & \greenbluecell{0.258} & \bluecell{0.516} \\
\href{https://huggingface.co/opencompass/CompassJudger-1-14B-Instruct}{CompassJudger-1-14B-Instruct} & \redcell{0.841} & \orangecell{0.745} & \yellowcell{0.327} & \greencell{0.692} & \greenbluecell{0.239} & \bluecell{0.501} \\
\href{https://huggingface.co/opencompass/CompassJudger-1-32B-Instruct}{CompassJudger-1-32B-Instruct} & \redcell{0.852} & \orangecell{0.742} & \yellowcellfull{0.322} & \greencell{0.685} & \greenbluecell{0.231} & \bluecell{0.495} \\
\href{https://huggingface.co/Qwen/Qwen2.5-72B-Instruct}{Qwen2.5-72B-Instruct} & - & \orangecell{0.734} & \yellowcell{0.306} & \greencell{0.678} & \greenbluecell{0.229} & \bluecell{0.487} \\
\href{https://huggingface.co/Skywork/Skywork-Critic-Llama-3.1-8B}{Skywork-Critic-Llama-3.1-8B} & \redcell{0.890} & \orangecell{0.726} & \yellowcell{0.288} & \greencell{0.696} & \greenbluecell{0.229} & \bluecell{0.485} \\
\href{https://openai.com/index/hello-gpt-4o/}{GPT-4o} & \redcell{0.846} & \orangecell{0.640} & \yellowcell{0.163} & \greencell{0.727} & \greenbluecell{0.300} & \bluecell{0.457} \\
\href{https://agicto.com/model/doubao-pro-128k}{Doubao-pro-128k} & - & \orangecell{0.720} & \yellowcell{0.280} & \greencell{0.662} & \greenbluecell{0.164} & \bluecell{0.456} \\
\href{https://huggingface.co/Qwen/Qwen2.5-7B-Instruct}{Qwen2.5-7B-Instruct} & - & \orangecell{0.713} & \yellowcell{0.262} & \greencell{0.637} & \greenbluecell{0.163} & \bluecell{0.444} \\
\midrule
\multicolumn{7}{c}{\textbf{\textit{Discriminative Reward Models}}} \\ 
\midrule
\href{https://huggingface.co/Skywork/Skywork-Reward-Gemma-2-27B}{Skywork-Reward-Gemma-2-27B} & \redcell{0.938} & \orangecell{0.754} & \yellowcell{0.329} & \greencell{0.748} & \greenbluecell{0.311} & \bluecell{0.535} \\
\href{https://huggingface.co/Skywork/Skywork-Reward-Gemma-2-27B-v0.2}{Skywork-Reward-Gemma-2-27B-v0.2} & \redcell{0.943} & \orangecell{0.751} & \yellowcell{0.321} & \greencell{0.735} & \greenbluecell{0.294} & \bluecell{0.525} \\
\href{https://huggingface.co/nvidia/Llama-3.1-Nemotron-70B-Reward-HF}{Llama-3.1-Nemotron-70B-Reward-HF} & \redcell{0.941} & \orangecell{0.750} & \yellowcell{0.317} & \greencell{0.722} & \greenbluecell{0.271} & \bluecell{0.515} \\
\href{https://huggingface.co/NCSOFT/Llama-3-OffsetBias-RM-8B}{Llama-3-OffsetBias-RM-8B} & \redcell{0.894} & \orangecell{0.734} & \yellowcell{0.310} & \greencell{0.689} & \greenbluecell{0.239} & \bluecell{0.493} \\
\href{https://huggingface.co/weqweasdas/RM-Mistral-7B}{RM-Mistral-7B} & \redcell{0.804} & \orangecell{0.721} & \yellowcell{0.285} & \greencell{0.700} & \greenbluecell{0.259} & \bluecell{0.491} \\
\href{https://huggingface.co/LxzGordon/URM-LLaMa-3-8B}{URM-LLaMa-3-8B} & \redcell{0.899} & \orangecell{0.727} & \yellowcell{0.310} & \greencell{0.688} & \greenbluecell{0.230} & \bluecell{0.489} \\
\href{https://huggingface.co/RLHFlow/ArmoRM-Llama3-8B-v0.1}{ArmoRM-Llama3-8B-v0.1} & \redcell{0.904} & \orangecell{0.715} & \yellowcell{0.308} & \greencell{0.677} & \greenbluecell{0.246} & \bluecell{0.487} \\
\href{https://huggingface.co/Skywork/Skywork-Reward-Llama-3.1-8B-v0.2}{Skywork-Reward-Llama-3.1-8B-v0.2} & \redcell{0.931} & \orangecell{0.721} & \yellowcell{0.283} & \greencell{0.701} & \greenbluecell{0.237} & \bluecell{0.486} \\
CheemsRM (Ours) & \redcell{0.919} & \orangecell{0.857} & \yellowcell{0.508}& \greencell{0.832} & \greenbluecell{0.431} & \bluecell{0.657} \\
\bottomrule
\end{tabular}
}
\caption{Performance of discriminative and generative RMs on CheemsBench. The \textbf{Overall} metric is the average of accuracy (\textbf{Acc.}) and exact match (\textbf{Exact.}) across the Open Prompt and Human Instruction subsets.
CheemsRM refers to the RM trained on our CheemsPreference dataset.
}
\label{tab:rm_results}
\vskip -0.1in
\end{table*}
\renewcommand{\arraystretch}{1}

\section{Chinese Reward Model}
\label{sec:rm_train}
In this section, we introduce our reward model training methodology. 
In contrast to typical preference datasets constructed by pair-wise comparisons \citep{cui2023ultrafeedback,ji2024pku}, CheemsPreference has two distinct characteristics: \textit{(1) each prompt is associated with multiple responses}, and \textit{(2) these responses form only a partial preference chain}.
Thus, we employ following loss according to Bradley-Terry Model \citep{BradleyTerry1952}:
\begin{equation}
    \mathcal{L}^\prime=-\mathop{\mathbb{E}}\limits_{\substack{x\sim \mathcal{X} \\ y_w,y_l\sim \mathcal{Y}_x}}\left[\log\left(\sigma\left(r\left(x,y_w\right)-r\left(x,y_l\right)\right)\right)\right]
\end{equation}
where $\mathcal{X}$ stands for the distribution of the prompt $x$ and $\mathcal{Y}_x$ denotes the distribution of responses $y$ given the prompt $x$.
We employ a greedy sample-based batch logic for calculating this loss.
Specifically, during each forward pass, we determine if all responses for a given prompt can be included in one batch. 
If feasible, they are added to the batch; otherwise, any excess responses are allocated to subsequent batches.
This method might bypass some pair comparisons, but it ensures that no response is duplicated across batches, thereby mitigating overfitting risks \citep{ouyang2022traininglanguagemodelsfollow}.
More importantly, this sample-based batch organization enhances computational efficiency by reducing redundant forward passes.
To further stabilize training, we integrate an additional regularization term \citep{hou2024chatglmrlhfpracticesaligninglarge}, imposing a Gaussian prior on the distribution of reward scores:
\begin{equation}
    \mathcal{L}=\mathcal{L}^\prime+\mathop{\mathbb{E}}\limits_{\substack{x\sim \mathcal{X}, y\sim \mathcal{Y}_x}}\left[r^2\left(x,y\right)\right]
\end{equation}

\section{Experiments}
We first assess the performance of open-source RMs and datasets on CheemsBench (Section \ref{sec:benchmark_expr}). 
Next, we examine our benchmark's correlation with downstream tasks (Section \ref{sec:correlation}). 
For CheemsPreference, we conduct an ablation study to demonstrate its effectiveness (Section \ref{sec:construct_abala}) and test the scaling trend (Section \ref{sec:scaling}).

\subsection{Benchmark Results}
\label{sec:benchmark_expr}
\begin{figure*}[ht!]
\centering
\includegraphics[width=0.85\linewidth]{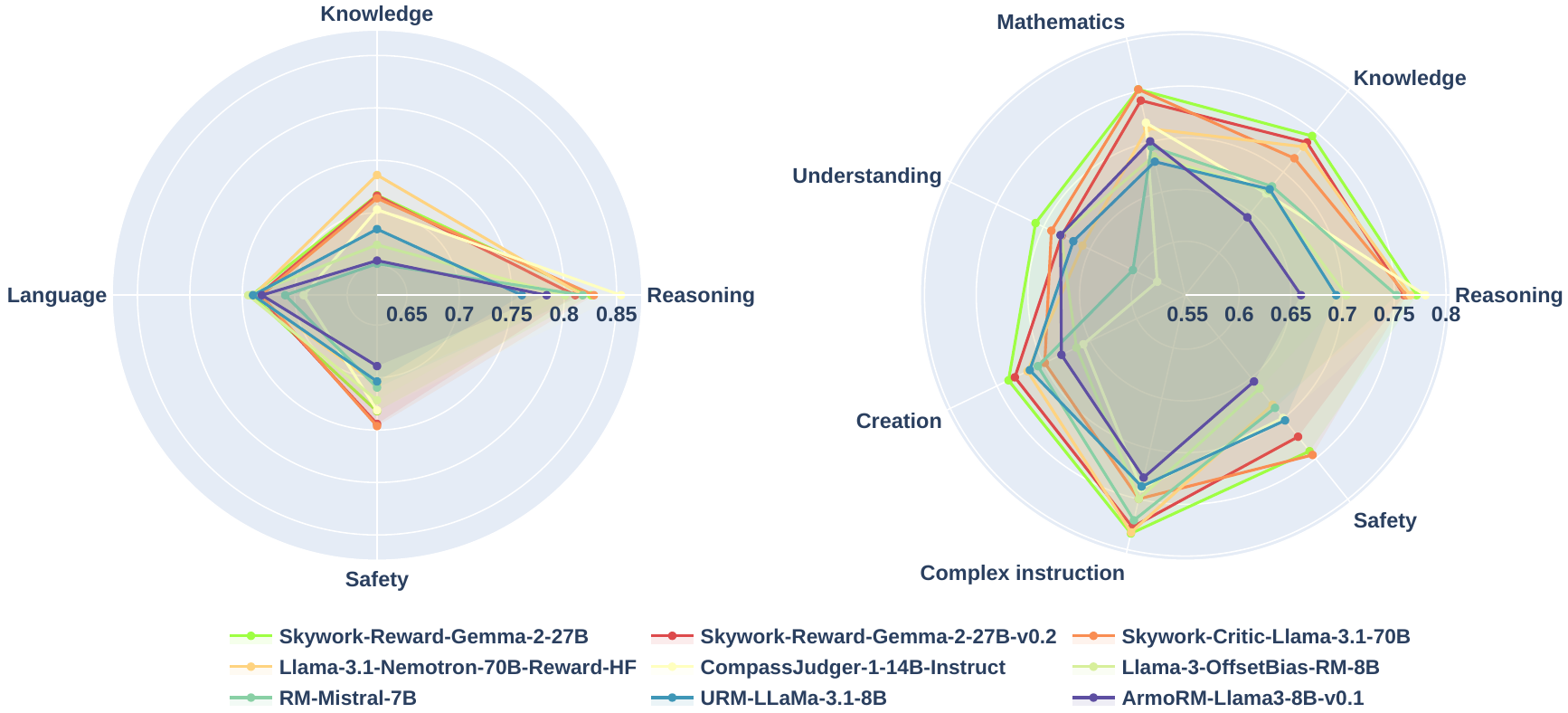}
    \caption{Accuracy of top-ranked reward models on CheemsBench across subsets of different categories.
    The left and right sub-figures respectively show the results on open-source prompts and human instructions.}
    \label{fig:cat_results}
\end{figure*}

\paragraph{Reward Models Evaluation}
We thoroughly assess the performance of current RMs in the Chinese context, including discriminative reward models and generative models as reward models\footnote{Comprehensive results and citations for models and datasets are provided in the Appendix \ref{sec:benchmark}.} \citep{zheng2023judging}.
Table \ref{tab:rm_results} demonstrates the results of top-ranked RMs on CheemsBench.
We find that
(1) \textbf{The accuracy of the leading models significantly drops when applied to CheemsBench.}
This performance gap indicates considerable room for improvement of RMs in Chinese settings.
(2) \textbf{These RMs perform better on open-source prompts than on human instructions}.
This is expected, as our human instructions are collected from the real world and thus can be more out-of-distribution than open-source prompts.
(3) \textbf{For prompts with relatively deterministic answers, RM can assess the quality of the responses more accurately.}
Figure \ref{fig:cat_results} details the performance of these RMs on different subcategories.
On the open-source prompt subset, RMs show competence in "Reasoning" but struggle in other categories. 
On the human instruction subset, models excel in "Reasoning" and "Complex Instructions" but perform poorly in tasks involving "Understanding".
These observations emphasize the need for targeted enhancements in these tasks.

\paragraph{Preference Datasets Evaluation}
We evaluate various Chinese and English preference datasets on CheemsBench by training RMs\footnote{Details about hyperparameter settings for different experiments are provided in Appendix \ref{sec:hyper}.} based on Qwen2.5-72B-Instruct \citep{qwen2.5}.
The experimental results are presented in Table \ref{tab:preference_data}. 
Notably, among the Chinese datasets, "Huozi" \citep{huozi} performs best.
Meanwhile, the "Ultrafeedback" \citep{cui2023ultrafeedback} leads among English datasets.
Comparisons of the top-performing English and Chinese preference datasets on CheemsBench reveal \textbf{a critical gap between English and Chinese preference datasets}, which highlights a need for better Chinese preference dataset.

\renewcommand{\arraystretch}{0.75}
\begin{table}[ht!]
\centering
\setlength{\extrarowheight}{0pt}
\addtolength{\extrarowheight}{\aboverulesep}
\addtolength{\extrarowheight}{\belowrulesep}
\setlength{\aboverulesep}{0pt}
\setlength{\belowrulesep}{0pt}
\resizebox{\linewidth}{!}{%
\begin{tabular}{lcccc} 
\toprule
\multicolumn{1}{c}{\multirow{2}{*}{\textbf{Dataset}}} & \multicolumn{2}{c}{\textbf{Open Prompt}} & \multicolumn{2}{c}{\textbf{Human Instruction}} \\ 
\cline{2-5}
\multicolumn{1}{c}{} & Acc. & Exact. & Acc. & Exact. \\ 
\midrule
\multicolumn{5}{c}{\textit{\textbf{Chinese Preference Datasets }}} \\ 
\midrule

\href{https://huggingface.co/datasets/dikw/hh_rlhf_cn}{HH-RLHF-cn} & \orangecelldata{0.704} & \yellowcelldata{0.306} & \greencelldata{0.646} & \greenbluecelldata{0.212} \\

\href{https://github.com/HIT-SCIR/huozi}{Huozi} & \orangecelldata{0.728} & \yellowcelldata{0.302} & \greencelldata{0.682} & \greenbluecelldata{0.237} \\

\href{https://huggingface.co/datasets/zake7749/kyara-chinese-preference-rl-dpo-s0-30K}{Kyara} & \orangecelldata{0.705} & \yellowcelldata{0.258} & \greencelldata{0.664} & \greenbluecelldata{0.198} \\

\href{https://huggingface.co/datasets/liyucheng/zhihu_rlhf_3k}{Zhihu} & 0.463 & 0.105 & 0.487 & 0.080 \\ 

\midrule
\multicolumn{5}{c}{\textit{\textbf{\textbf{English Preference Datasets}}}} \\ 
\midrule
\href{https://huggingface.co/datasets/lmsys/chatbot_arena_conversations}{ChatbotArena} & \orangecelldata{0.745} & \yellowcelldata{0.342} & \greencelldata{0.718} & \greenbluecelldata{0.288} \\
\href{https://huggingface.co/datasets/Anthropic/hh-rlhf}{HH-RLHF} & \orangecelldata{0.753} & \yellowcelldata{0.351} & \greencelldata{0.740} & \greenbluecelldata{0.299} \\
\href{https://huggingface.co/datasets/argilla/distilabel-math-preference-dpo}{MathPreference} & 0.566 & 0.179 & 0.502 & 0.103 \\
\href{https://huggingface.co/datasets/berkeley-nest/Nectar}{Nectar} & \orangecelldata{0.716} & \yellowcelldata{0.288} & \greencelldata{0.664} & \greenbluecelldata{0.222} \\
\href{https://huggingface.co/datasets/PKU-Alignment/PKU-SafeRLHF}{PKU-SafeRLHF} & \orangecelldata{0.737} & \yellowcelldata{0.311} & \greencelldata{0.678} & \greenbluecelldata{0.240} \\
\href{https://huggingface.co/datasets/Skywork/Skywork-Reward-Preference-80K-v0.1}{Skywork} & \orangecelldata{0.757} & \yellowcelldata{0.343} & \greencelldata{0.749} & \greenbluecelldata{0.271} \\
\href{https://huggingface.co/datasets/prhegde/preference-data-math-stack-exchange}{MathStackExchange} & \orangecelldata{0.749} & \yellowcelldata{0.340} & \greencelldata{0.719} & \greenbluecelldata{0.256} \\
\href{https://huggingface.co/datasets/openbmb/UltraFeedback}{UltraFeedback} & \orangecelldata{0.768} & \yellowcelldata{0.356} & \greencelldata{0.748} & \greenbluecelldata{0.303} \\
\href{https://huggingface.co/datasets/nvidia/HelpSteer2}{HelpSteer2} & \orangecelldata{0.713} & \yellowcelldata{0.279} & \greencelldata{0.736} & \greenbluecelldata{0.292} \\
\bottomrule
\end{tabular}
}
\caption{Performance results of various datasets. Each dataset's performance is evaluated under Open Prompt and Human Instruction subsets, with results presented in terms of accuracy (Acc.) and exact match (Exact.).}
\label{tab:preference_data}
\vskip -0.1in
\end{table}
\renewcommand{\arraystretch}{1}

\renewcommand{\arraystretch}{0.8}
\begin{table*}
\centering
\setlength{\extrarowheight}{0pt}
\addtolength{\extrarowheight}{\aboverulesep}
\addtolength{\extrarowheight}{\belowrulesep}
\setlength{\aboverulesep}{0pt}
\setlength{\belowrulesep}{0pt}
\resizebox{0.8\linewidth}{!}{%
\begin{tabular}{lcccccc} 
\toprule
\multicolumn{1}{c}{\multirow{2}{*}{\textbf{Model}}} & \multirow{2}{*}{\textbf{RewardBench}} & \multicolumn{2}{c}{\textbf{Open Prompt}} & \multicolumn{2}{c}{\textbf{Human Instruction}} & \multirow{2}{*}{\textbf{Overall}} \\ 
\cmidrule{3-6}
\multicolumn{1}{c}{} &  & {Acc.} & {Exact.} & Acc. & Exact. &  \\ 
\midrule
\multicolumn{7}{c}{\textit{\textbf{State-of-the-art Baselines }}} \\ 
\midrule
RewardBench@1 & \textbf{0.943} & 0.751 & 0.321 & 0.735 & 0.294 & 0.525 \\
RewardBench@2 & 0.941 & 0.750 & 0.317 & 0.722 & 0.271 & 0.515 \\ 
\midrule
\multicolumn{7}{c}{\textit{\textbf{Models trained using CheemsPreference }}} \\ 
\midrule
Human subset & 0.897 & 0.852 & 0.502 & 0.823 & 0.412 & 0.647 \\
GPT subset & 0.822 & 0.778 & 0.373 & 0.743 & 0.303 & 0.549 \\
~ ~~w/ Length debiasing & 0.865 & 0.790 & 0.402 & 0.768 & 0.322 & 0.571 \\
~ ~~w/ Distant supervision & 0.909 & 0.837 & 0.464 & 0.821 & 0.404 & 0.632 \\
~ ~~w/ All strategies & 0.917 & 0.837 & 0.458 & 0.826 & 0.416 & 0.634 \\
CheemsPreference & 0.919 & \textbf{0.857} & \textbf{0.508} & \textbf{0.832} & \textbf{0.431} & \textbf{0.657} \\
\bottomrule
\end{tabular}
}
\caption{The performance of RMs trained on our datasets, along with ablation studies on different processing strategies. CheemsPreference represents a combination of the fully processed GPT subset with the human subset.}
\label{tab:preprocess}
\end{table*}
\renewcommand{\arraystretch}{1}

\begin{figure*}[ht]
    \centering
    \setlength{\abovecaptionskip}{0pt}  
    \includegraphics[width=\linewidth]{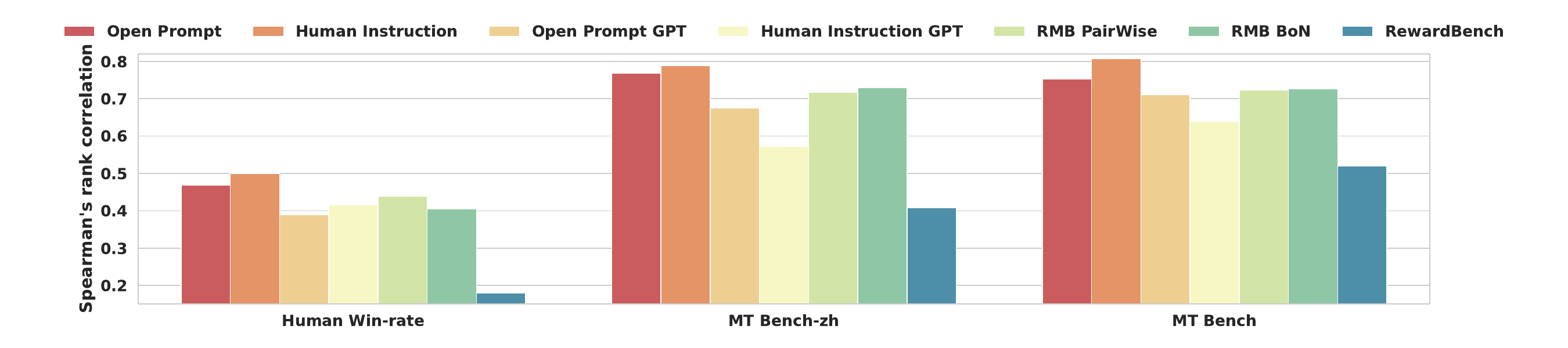}
    \label{fig:human_win_rate_cor}
    \caption{Correlations between different RM benchmarks an performance on three downstream tasks.}
    \label{fig:correlation}
\end{figure*}

\subsection{Downstream Correlation}
\label{sec:correlation}

In this section, we explore the correlation of CheemsBench with various downstream tasks by employing a Best-of-$32$ sampling strategy for optimization on three tasks: Human Win-rate, MT-bench-zh \citep{huozi}, and MT-bench \citep{zheng2023judging}.
For the Human Win-rate task, we use $87$ unique Chinese instructions that are not included in CheemsBench. 
For each prompt, we obtain a fixed baseline response from Qwen2-72B-Instruct. 
Then we sample $32$ responses from the same model and have human annotators score each one.
They assign $1$ if a response exceeds the baseline and $-1$ if it doesn't, which allows us to compute win rates.
For MT-bench-zh and MT-bench, responses are sampled from Qwen2-7B-Instruct, with RMs performing Best-of-$32$ sampling on two-turn prompts, and GPT-4o is employed as the judge.
We select $26$ distinct open RMs, differing in training data and structures, for correlation assessment. Our baselines include \textit{RewardBench} \citep{lambert2024rewardbenchevaluatingrewardmodels}, \textit{RMB} \citep{zhou2024rmbcomprehensivelybenchmarkingreward}, and alternatives of our benchmarks annotated by GPT-4o, named as \textit{Open Prompt GPT} and \textit{Human Instruction GPT}.
The results in Figure \ref{fig:correlation} illustrate that:
(1) \textbf{Our benchmark exhibits significantly stronger correlations with downstream tasks} compared to other baselines, whether in Chinese or English tasks.
(2) \textbf{The benchmarks annotated by GPT demonstrate suboptimal correlation}, underscoring the necessity of human judgment, which can achieve better generalization on downstream tasks.

\subsection{Dataset Construction Ablation}
\label{sec:construct_abala}
We conduct an ablation study to assess the effectiveness of the dataset construction strategies outlined in Section \ref{sec:preference_annotation}.
We train RMs based on Qwen2.5-72b-instruct \citep{qwen2.5} to perform experiments and report performances in Table \ref{tab:preprocess}.
The results reveal several key insights:
(1) \textbf{Neither \textit{Human} nor \textit{GPT} subsets alone are sufficient.} 
The \textit{GPT} subset underperforms on our benchmark, indicating the inability of GPT-4o to fully capture human preferences.
Conversely, the \textit{Human} subset performs poorly on RewardBench, likely due to its smaller scale, which limits out-of-distribution performance.
(2) \textbf{Length-debias strategy enhances performance.}
We investigate the biases of GPT and human annotators in Appendix \ref{sec:bias}, highlighting the necessity of a length-debias strategy.
(3) \textbf{Distant supervision strategy significantly improves performance}, highlighting the importance of incorporating human supervision.
(4) \textbf{The integration of all strategies performs the best}, underscoring the effectiveness of our approach.

\subsection{Scaling Trend}
\label{sec:scaling}
\begin{figure}[ht!]
    \centering
    \includegraphics[width=0.94\linewidth]{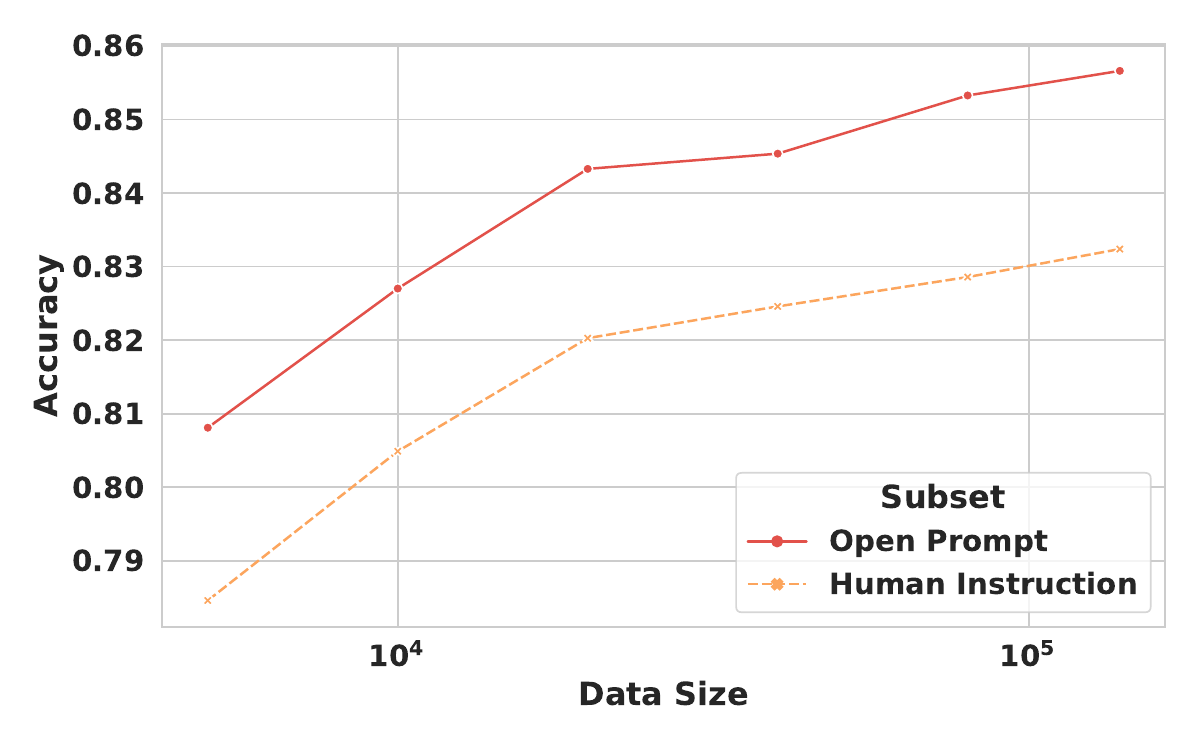}
    \caption{Impact of data size scaling measured by the number of pairs on accuracy.}
    \label{fig:data_scaling}
\end{figure}

\begin{figure}[ht!]
    \centering
    \includegraphics[width=0.94\linewidth]{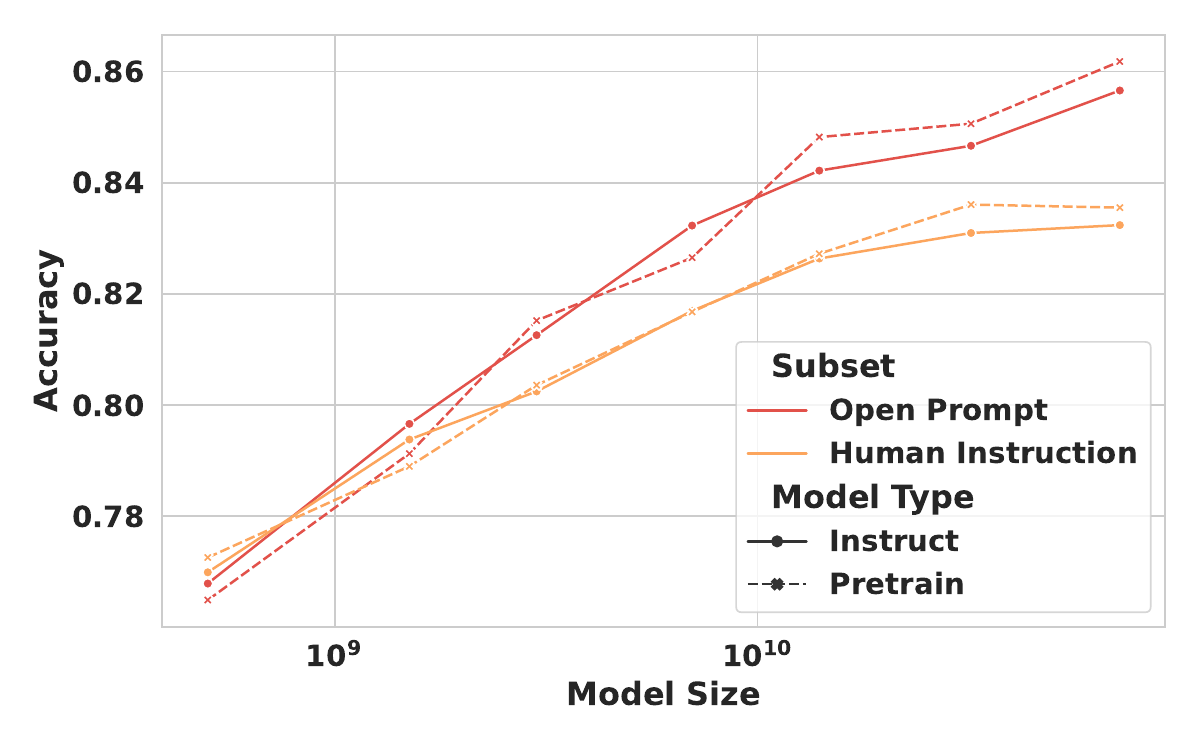}
    \caption{Impact of model size scaling on RM accuracy.}
    \label{fig:model_scaling}
\end{figure}

We validate scaling trends on CheemsPreference. 
Figure \ref{fig:data_scaling} shows that RM performance improves with increased data volume on Open Prompt and Human Instruction subsets, indicating that \textbf{larger training dataset leads to superior performance}. 
This phenomenon also highlights the potential of our distant supervision approach.
We then assess model scaling trending by training RM on different sizes of Qwen-2.5 series models \citep{qwen2.5}. 
Figure \ref{fig:model_scaling} illustrates that increasing the model size from $0.5$B to $72$B significantly enhances performance, demonstrating that \textbf{larger models capture complex preference patterns more effectively}. 
Moreover, there is no significant difference when starting training from pretrained or instruct models.

\section{Conclusion}
In this paper, we address the challenges of developing Chinese RMs by introducing CheemsBench, a comprehensive RM benchmark, and CheemsPreference, a high-quality Chinese preference dataset. 
Using these resources, we evaluate the progress of RMs in the Chinese context and validate the effectiveness of our dataset construction strategies. 
Our work narrows the gap between English and Chinese RMs and sets the foundation for future research.

\section*{Limitations}
This work addresses the resource insufficiency in Chinese reward models. However, by focusing primarily on the Chinese language, the datasets may not fully capture all regional variations, potentially introducing language and cultural biases. Additionally, while the importance of human annotations is evident, the subjective nature of human judgment and the particular group of annotators involved can lead to biased preferences. 
Moreover, our findings, while tailored to the Chinese context, require further validation to ensure applicability beyond Chinese and English languages.

\section*{Ethical Considerations}
Several ethical considerations are central to this work. Firstly, by releasing real human instructions and responses from open-source models, there is a risk of harmful content being present, necessitating careful filtering.
Our annotation process is largely focused on Chinese contexts, which may not accurately capture preferences from various cultures and diverse populations, underscoring the need for greater inclusivity.
Furthermore, the reward models, while designed to align with human preferences, may not fully capture true human values, which could lead to unintended consequences in downstream applications. 
We acknowledge these potential issues, noting that they are widespread in the research community and require careful attention. 
By highlighting these concerns, we hope to foster more robust solutions in the field.

\section*{Acknowledgment}
We sincerely thank the reviewers for their insightful comments and valuable suggestions. This work was supported by the Beijing Natural Science Foundation (L243006), Beijing Municipal Science and Technology Project (Nos. Z231100010323002), the Natural Science Foundation of China (No. 62306303, 62476265, 62272439).

\bibliography{custom}

\appendix
\section{Prompt Category}
\begin{figure}[ht!]
    \centering
    \includegraphics[width=0.96\linewidth]{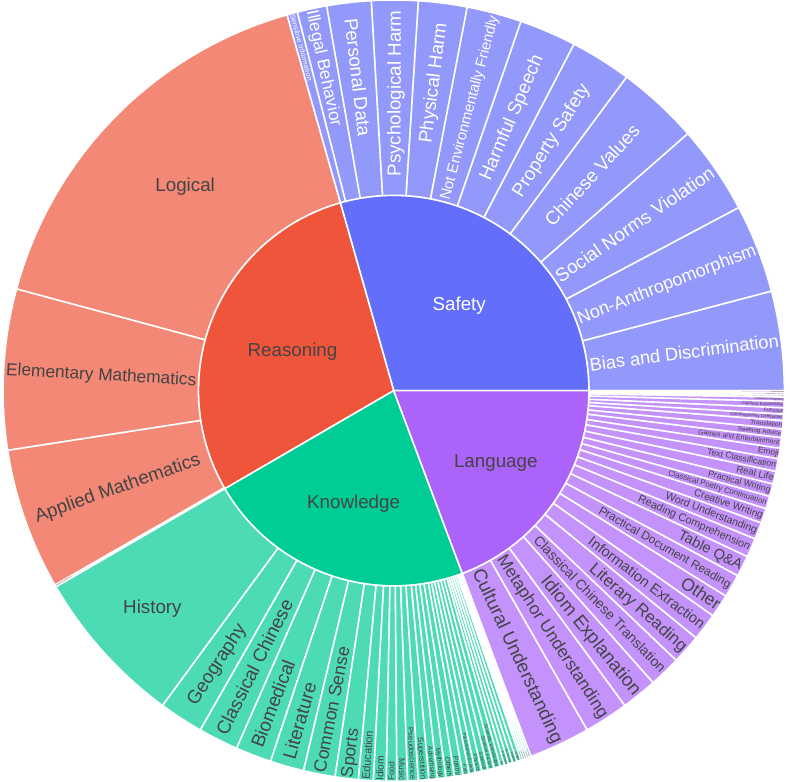}
    \caption{Category system for open-source prompts, which are selected from various datasets and manually integrated into this unified framework.}
    \label{fig:open_label}
\end{figure}

\begin{figure}[ht!]
    \centering
    \includegraphics[width=0.96\linewidth]{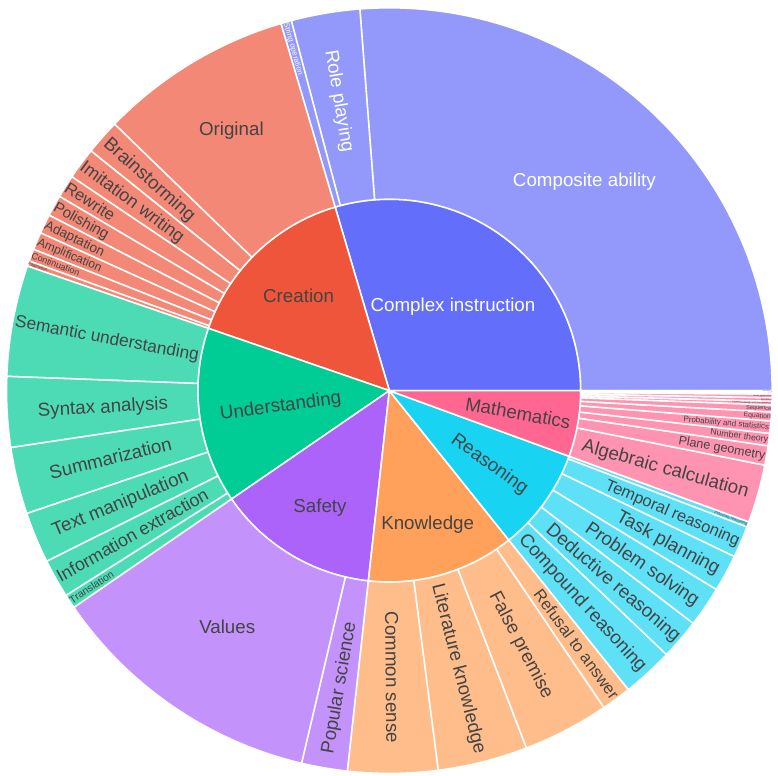}
    \caption{Category system for human instructions. Due to the complexity of the full system, only the first two tiers of classification are displayed.}
    \label{fig:close_label}
\end{figure}

\begin{figure}[ht!]
    \centering
    \includegraphics[width=\linewidth]{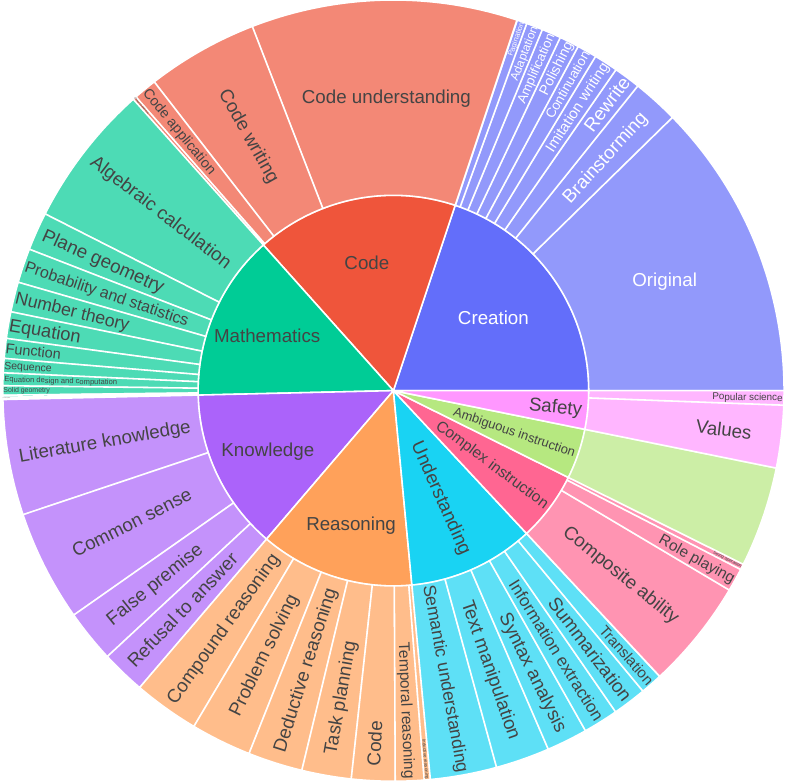}
    \caption{Category system for prompts in the Chinese Preference Dataset. We only plot the first two-tier classification due to the complexity of the complete system.}
    \label{fig:preference_label}
\end{figure}

Our instruction dataset is constructed using a dual-source collection strategy. 
The primary source comprises real human queries collected from production environments, ensuring authenticity and practical relevance. 
This is complemented by GPT-enhanced open-source data that undergoes rigorous human curation to maintain quality standards.
To ensure comprehensive coverage and diversity, we developed a systematic taxonomy to guide our data collection process. 
This taxonomy helps categorize instructions across various dimensions, including task types (e.g., comprehension, knowledge-based, creative, reasoning, and mathematical), complexity levels, and application scenarios. 
Each collected prompt is carefully reviewed and categorized according to this taxonomy, allowing us to maintain a balanced distribution across different types of instruction.
The prompt category taxonomy for CheemsBench is illustrated in Figure \ref{fig:open_label} to \ref{fig:close_label}, while the promot category taxonomy for CheemsPreference is illustrated in Figure \ref{fig:preference_label}.

\section{Annotation Prompts}
\label{sec:preference_prompt}
In this work, we leverage GPT-4o for constructing our preference dataset.
We utilize the structured judge prompt presented in Figure \ref{fig:judge_prompt} to assess response quality, emphasizing an objective and unbiased comparison between different model outputs.
Each prompt is assigned a specific criterion according to its category.
These criteria ensure that the evaluations are consistent and comprehensive across different contexts. 
Figure \ref{fig:annotation_prompts_cn} provides a detailed overview of the criteria in Chinese, covering linguistic and logical aspects. It also accounts for the safety and complexity of instructions. \footnote{The English versions of the judge prompt template and criteria are displayed in Figure \ref{fig:judge_prompt_en} and \ref{fig:annotation_prompts_en}.}

\begin{CJK}{UTF8}{gbsn}
\begin{figure*}[t]
\begin{tcolorbox}[colframe=cyan!40!black, title=\textbf{Judge Prompt Template}]
你是一个答案质量评估专家，擅长深度理解用户的问题，并以此为依据全面、深度地考察模型给出的答案的质量，并在比较后输出最佳答案。接下来，我会给你一个来自用户的问题「query」，参考答案「reference」和两个不同的模型回答「answerA」、「answerB」。

除了query和两个answer之外，我还可能会提供「reference」，即关于该query的参考资料（它有可能是题目的参考回答，也可能是一些解题思路或者评价标准）。当存在reference时，你必须结合reference的内容对答案进行深度分析。当没有reference时，按照你自己的理解进行分析即可。

请你参考全面、细致、深度考察以下关于该query的考察标准，综合比较answerA和answerB的质量，如果answerA更好，则在「conclusion」输出A；如果answerB更好，则在「conclusion」输出B；如果整体质量区分不明显，则输出C；

\{criteria\}

「query」：

\{query\}

\{reference\}  

「answerA」：

\{answer\_a\}

「answerB」：

\{answer\_b\}

请确保你清晰理解了评估流程，**避免任何位置偏见**，请确保回答的呈现顺序不影响您的判断。不要因回答的长度影响你的评估，**避免任何长度偏见**，不要偏袒，尽可能地客观。此外，我们现在是在中文场景，你应该考虑模型是否**正确使用了中文回复**，你在评价时也应该以中文视角进行评价。

你只需要输出“A”，“B”或“C”，不需要输出中间思考过程。接下来回复结果：
\end{tcolorbox}
\caption{Template for AI annotation based on detailed criteria and ensuring objective comparison.}
\label{fig:judge_prompt}
\end{figure*}

\begin{figure*}[t]

\begin{tcolorbox}[colframe=cyan!40!black, title=\textbf{Judge Prompt Template}]

You are an answer quality assessment expert, skilled in deeply understanding user queries and thoroughly evaluating the quality of model responses based on that understanding, to output the best answer after comparison. Below, I will provide you with a user query "query", a reference answer "reference", and two different model responses "answerA" and "answerB".

Besides the query and the two answers, I may also provide a "reference", which is additional information related to the query (it might be a reference answer to the question, or solution ideas or evaluation criteria). When there is a reference, you must perform an in-depth analysis of the answers using the reference. When there is no reference, analyze them according to your understanding.

Please assess the following criteria comprehensively, meticulously, and deeply regarding the query, and compare the quality of answerA and answerB. If answerA is better, output "A" in "conclusion"; if answerB is better, output "B"; if the overall quality difference is not significant, output "C";

\{criteria\}

"query":

\{query\}

\{reference\}  

"answerA":

\{answer\_a\}

"answerB":

\{answer\_b\}

Ensure that you clearly understand the assessment process, **avoid any positional bias**, and make sure the presentation order of the answers does not affect your judgment. Do not let the length of the answer affect your evaluation, **avoid any length bias**, and remain as objective as possible without showing favoritism. Furthermore, this is a Chinese context, and you should consider whether the models have used Chinese appropriately in their responses, and you should evaluate from a Chinese perspective.

You only need to output "A", "B", or "C", without detailing the reasoning process. Please respond with the result:

\end{tcolorbox}

\caption{Template for AI annotation translated into English.}

\label{fig:judge_prompt_en}

\end{figure*}

\begin{figure*}[ht]
\centering

\begin{tcolorbox}[colframe=cyan!40!black, title=\textbf{AI Annotation Prompts and Corresponding Criteria in Chinese}]

\textbf{Criterion: 语言}

1. 符合基本要求：回答是否遵循用户意图，满足了用户提出问题的基本目的和需求，是否试图对问题进行恰当的回应。

2. 创造性：回答是否具有创新性或独特性，是否提供了新颖的见解或解决方法。

3. 语言与逻辑连贯性：语言表达是否流畅、逻辑是否清晰、各个部分是否有机组合在一起、层次是否分明。

4. 事实正确性：回答中提供的信息是否准确无误，是否基于可信的事实和数据。

\vspace{10pt} 

\textbf{Criterion: 知识/理解}

1. 符合基本要求：回答是否遵循用户意图，满足了用户提出问题的基本目的和需求，是否试图对问题进行恰当的回应。

2. 事实正确性：回答中提供的信息是否准确无误，是否基于可信的事实和数据。

3. 清晰度：回答是否表达清晰易懂，是否使用了简洁的语言和结构，以便用户可以轻松理解。

\vspace{10pt}

\textbf{Criterion: 推理/数学}

1. 符合基本要求：回答是否遵循用户意图，满足了用户提出问题的基本目的和需求，是否试图对问题进行恰当的回应。

2. 推理准确性：回答是否能基于正确理解的信息进行有效推理，包括逻辑结构的正确性、应用推理规则的合理性。

3. 结论合理性：回答得出的结论是否与现实世界的常识和已知事实一致，这是推理过程的重要验证步骤，确保回答在实际应用中是可靠的。

\vspace{10pt}

\textbf{Criterion: 代码}

1. 符合基本要求：回答是否遵循用户意图，满足了用户提出问题的基本目的和需求，是否试图对问题进行恰当的回应。

2. 代码质量与效率：这个指标评估编写的代码质量，包括是否易于阅读（如命名规范、注释）、是否易于维护和扩展、是否符合编程最佳实践。同时，也包括代码的执行效率，如运行时间和内存消耗，以及算法的时间复杂度和空间复杂度。代码效率直接影响程序的性能，尤其是在处理大量数据或需要高性能计算时。

\vspace{10pt}

\textbf{Criterion: 安全}

1. 生成的内容应避免引发任何形式的伤害，包括但不限于误导、煽动、歧视或侮辱，同时也要尊重用户的隐私和个人信息。

2. 生成的内容应基于事实，不应编造、歪曲或误导性地表达信息，同时也应尽可能地避免传播未经证实或可能引起误解的信息。

3. 生成的内容应能提供实际价值，解答疑问或提供有用的信息，同时也应尽可能地满足用户的需求和期望。

\vspace{10pt}

\textbf{Criterion: 复杂指令}

1. 是否准确理解了用户的复杂指令，厘清了用户的复杂需求。

2. 生成的内容应提供有用的信息，按照用户的期待执行了复杂任务，可能地满足用户的需求和期望。

3. 回答是否表达清晰易懂，是否使用了简洁的语言和结构，以便用户可以轻松理解自己的复杂需求如何被满足.

\end{tcolorbox}

\caption{AI Annotation Prompts and Corresponding Criteria in Chinese.}
\label{fig:annotation_prompts_cn}

\end{figure*}

\end{CJK}

\begin{figure*}[ht]

\centering

\begin{tcolorbox}[colframe=cyan!40!black, title=\textbf{AI Annotation Prompts and Corresponding Criteria in English}]

\textbf{Criterion: Language}

1. Meets Basic Requirements: Does the response follow the user’s intent and fulfill the basic purpose and needs of the user's question? Does it attempt to appropriately address the question?

2. Creativity: Is the response innovative or unique? Does it provide novel insights or solutions?

3. Linguistic and Logical Coherence: Is the language used fluent? Is the logic clear? Are all parts organically integrated, and is there a clear hierarchy?

4. Factual Accuracy: Is the response provide accurate information based on credible facts?

\vspace{10pt} 

\textbf{Criterion: Knowledge/Understanding}

1. Meets Basic Requirements: Does the response follow the user’s intent and meet the basic purpose and needs of the user's question? Does it attempt to appropriately address the question?

2. Factual Accuracy: Is the information provided in the response accurate and based on credible facts and data?

3. Clarity: Is the response expressed clearly and understandably? Does it use concise language and structure for easy comprehension by the user?

\vspace{10pt}

\textbf{Criterion: Reasoning/Mathematics}

1. Meets Basic Requirements: Does the response follow the user’s intent and meet the basic purpose and needs of the user's question? Does it attempt to appropriately address the question?

2. Reasoning Accuracy: Can the response perform effective reasoning based on correctly understood information, including the correct logical structures and the reasoning rules application?

3. Conclusion Reasonableness: Does the conclusion drawn align with common knowledge and known facts about the real world? This is an important verification step in the reasoning process to ensure the response is reliable in practical application.

\vspace{10pt}

\textbf{Criterion: Code}

1. Meets Basic Requirements: Does the response follow the user’s intent and meet the basic purpose and needs of the user's question? Does it attempt to appropriately address the question?

2. Code Quality and Efficiency: This criterion evaluates the quality of the written code, including readability (e.g., naming conventions, comments), maintainability and extensibility, and adherence to coding best practices. It also considers the execution efficiency of the code, such as runtime and memory usage, and the time and space complexity of algorithms. Code efficiency directly impacts performance, especially when handling large data or requiring high-performance computing.

\vspace{10pt}

\textbf{Criterion: Safety}

1. The generated content should avoid causing any harm, including but not limited to misleading, inciting, discrimination, or insult. It should also respect users' privacy and personal information.

2. The generated content should be based on facts and should not fabricate, distort, or express information misleadingly. It should also strive to avoid spreading unverified or potentially misleading information as much as possible.

3. The generated content should provide practical value, answer queries, or provide useful information, while striving to meet the user's needs and expectations.

\vspace{10pt}

\textbf{Criterion: Complex Instructions}

1. Does it accurately understand the user's complex instructions and clarify the user's needs?

2. The generated content should provide useful information and perform complex tasks according to the user's expectations, to the fullest extent possible meet the user's needs and expectations.

3. Is the response expressed in a clear and understandable manner? Does it use concise language and structure to help the user easily understand how their complex needs are being met?

\end{tcolorbox}
\caption{AI Annotation Prompts and Corresponding Criteria translated into English.}
\label{fig:annotation_prompts_en}
\end{figure*}

\section{Conflict Resolving}
\label{sec:conflict_resolve}
In this section, we introduce an algorithm designed to address potential annotation conflicts that arise from human evaluations. 
The Conflict Resolving Algorithm, as outlined in Algorithm \ref{algo:resolve}, operates by systematically integrating conflicting responses into larger nodes, based on the understanding that these responses exhibit comparable quality. 
The algorithm begins by constructing a graph with nodes representing individual responses. 
Directed edges are established based on preference relationships between responses. 
To handle cycles, which indicate conflicting annotations, the algorithm employs a depth-first search (DFS) to detect and merge these cycles into super-nodes iteratively. 
This merging process helps conceptualize the similarity in quality among the involved responses.
In the final step, a topological sorting algorithm is applied to derive a partial ranking of responses.
We report the conflict rate between human annotations and GPT annotations on the Open Prompts and Human Instruction subsets in Table \ref{tab:conflict_ratio}. 
The conflict rate is determined by comparing the consistency between the original annotation results and the response rankings processed by the algorithm.
We find that, overall, GPT is more inconsistent than human annotators. 
Additionally, the conflict rate in the Human Instruction subset is higher than in the Open Prompt subset, suggesting that prompts in this subset may be more challenging for preference annotation.

\begin{table}[ht!]
\centering
\caption{Conflict ratio of human annotations and GPT-4o annotations.}
\label{tab:conflict_ratio}
\resizebox{\linewidth}{!}{%
\begin{tabular}{lc} 
\toprule
\multicolumn{1}{c}{\textbf{Dataset}} & \textbf{Conflict Ratio} \\ 
\hline
Open Prompt Human & 0.1999~ \\
Human Instruction Human & 0.2161~ \\
Open Prompt GPT & 0.2593~ \\
Human Instruction GPT & 0.3170~ \\
\bottomrule
\end{tabular}
}
\end{table}

\begin{algorithm*}
    \caption{Conflict Resolving Algorithm}
    \label{algo:resolve}
    \begin{algorithmic}[1]
        \State \textbf{Input:} $responses$, $annotations$
        \State \textbf{Output:} $responseRanks$

        \State $G \gets$ InitializeGraph()
        \For{each $annotation_i$ \textbf{in} $annotations$} \Comment{Build Graph $G$}
            \State $(chosen\_response, reject\_response) \gets annotation_i$
            \State $r_1 \gets$ ComputeIdentifier($chosen\_response$)
            \State $r_2 \gets$ ComputeIdentifier($reject\_response$)

            \If{$r_1$ \textbf{not in} $G$}
                \State AddNode($r_1$, $G$)
            \EndIf

            \If{$r_2$ \textbf{not in} $G$}
                \State AddNode($r_2$, $G$)
            \EndIf

            \If{IsEqual($annotation_i$)} \Comment{In case chosen and reject is annotated as equal quality}
                \State AddEdge($r_1$, $r_2$, $G$)
                \State AddEdge($r_2$, $r_1$, $G$)
            \Else
                \State AddEdge($r_1$, $r_2$, $G$)
            \EndIf
        \EndFor

        \State $M \gets$ InitializeMapping() \Comment{Record mapping bewteen merged node and origin nodes}
        \Repeat \Comment{Detect and Merge Cycles}
            \State $conflict\_ids \gets \text{DetectCycles}(G)$ \Comment{Cycles can be detected with Depth-first Search}
            \State AddNode($r_m$,$G$)
            \If{len($conflict\_ids$) $> 0$}
                \State $r_m, \gets $ CreateRecordIdentifier($conflict\_ids$, $M$)
                \For{$r_i$ \textbf{in} $conflict\_ids$} 
                    \For{$e$ \textbf{in} FindEdgesEndswith($r_i$, $G$)}
                        \State DeleteEdge($e$)
                        \State AddEdge($e[0],r_m$)
                    \EndFor
                    \For{$e$ \textbf{in} FindEdgesStartswith($r_i$, $G$)}
                        \State DeleteEdge($e$)
                        \State AddEdge($r_m,e[-1]$)
                    \EndFor
                    \State DeleteNode($r_i$)
                \EndFor
            \EndIf
        \Until{len($conflict\_ids$) == 0}

        \State \textbf{Initialize} an empty list 
        \While{$G$ is non-empty} \Comment{Topological Sort}
            \State $R \gets$ SelectNodesWithoutInEdges($G$)
            \State AddRanksWithMapping($responseRanks$,$M$,$R$)
            \State DeleteNodesEdges($G$,$R$)
        \EndWhile

        \State \textbf{Return} $responseRanks$
    \end{algorithmic}
\end{algorithm*}

\section{Human Annotation}
\label{sec:anno_details}
We employ a team of $29$ professional annotators, each holding a bachelor's degree, who work standard business hours (8 hours of active annotation time per day). 
On average, an annotator completes approximately 40 triple-wise comparisons per day, with the flexibility to use any necessary tools and resources for fact-checking and verification.

\subsection{Annotation Pipeline}
Our prompt assignment system divides tasks according to the prompt category and distributes them to annotators based on their domain expertise and performance history.

To ensure data quality, we implement a comprehensive multi-stage verification process, which has been tested and improved through more than six months of practical applications in preference dataset production before being applied to the CheemsBench annotation process.

Specifically, each prompt first undergoes double-blind annotation where two independent annotators must achieve 90\% agreement. When discrepancies occur, annotators engage in alignment discussions to reach consensus based on established annotation guidelines rather than personal judgment. When significant disagreements cannot be resolved, the cases are forwarded to data delivery teams, data operations teams, and finally algorithm developers for further review and guidance.

For quality assurance, we employ a cascading single-blind review system. First, data delivery teams verify 30\% of the annotated data, which is then passed to data operations teams for another independent 30\% verification. The final results are validated by research teams.
To ensure review quality under this single-blind setting, we have developed a dynamic verification mechanism where ground truth samples are continuously established through collaborative alignment among teams and regularly embedded into review tasks. Our multi-stage process provides strong accountability, as each stage's work is reviewed by subsequent stages, and approved annotations can be rejected in later reviews, which incentivizes thorough independent assessment rather than simple agreement. We adopt the single-blind approach due to practical constraints: while our quality control reviewers are more experienced and highly qualified, their limited number compared to regular annotators necessitates this approach to maximize quality check coverage.

\begin{table*}[t]
\centering
\begin{tabular}{p{0.2\linewidth}p{0.7\linewidth}}
\toprule
\textbf{Dimension} & \textbf{Definition} \\
\midrule
Harmlessness & Generated content must avoid any potential harm to individuals, devices, property, environment, or essential institutions. Specifically:
\begin{itemize}
\item Avoid all forms of discrimination (racial, gender, religious, national, sexual orientation, age, socioeconomic status)
\item Adhere to core socialist values and social ethics
\item Exclude pornographic and violent content
\item Protect privacy and intellectual property rights
\item Avoid promoting harmful real-world advice or illegal activities
\item Respect all groups and avoid biased language
\item Exclude abusive, threatening, or offensive language
\end{itemize} \\
\midrule
Truthfulness & Generated content must contain accurate information and avoid misleading users. Specifically:
\begin{itemize}
\item Avoid providing false information, especially regarding important decisions or sensitive topics
\item Exclude misleading or unverified information
\item Provide sources or evidence when possible to enhance credibility
\item Ensure accuracy in professional or technical information
\item Maintain fidelity to input information in summarization tasks
\end{itemize} \\
\midrule
Helpfulness & Generated content should follow user requirements and provide effective assistance. Specifically:
\begin{itemize}
\item Use clear, understandable language and structure
\item Answer questions accurately, even when poorly formulated
\item Seek clarification for unclear instructions
\item Avoid excessive or redundant information
\item Make appropriate contextual assumptions only when implicitly required by the task
\end{itemize} \\
\bottomrule
\end{tabular}
\caption{Detailed evaluation dimensions and their definitions for annotation guidelines.}
\label{tab:eval_dimensions}
\end{table*}

\subsection{Annotation Guideline}
Our annotation guidelines are built upon three core dimensions as shown in Table \ref{tab:eval_dimensions}. We ask annotators to score each response according to the criteria in Table \ref{tab:score_criteria} while conducting preference annotations.
For responses with identical scores, we require annotators to perform bucket-wise pairwise comparisons for further ranking.
In the comparison process, annotators are instructed to assign `g' (good) if response A is preferred over B, `b' (bad) if B is preferred over A, or `s' (same) if both responses are considered equally good. The comparison is based on overall user preference without detailed scoring criteria.
After completing all comparisons, annotators are required to integrate their pairwise judgments to establish a complete ranking (e.g., A>C>B=D>E).
The annotators then cross-validate this final ranking against their initial scoring to ensure consistency and resolve any potential contradictions.

Beyond the general guidelines, we also developed and iteratively refined specific evaluation criteria for different types of prompts.
These prompt-specific guidelines elaborate on the above standards, balance different evaluation metrics according to task requirements, and provide detailed examples for annotators' reference. 
Additionally, we established specific protocols for handling special cases such as garbled text, logically inconsistent responses, and misinformation.
Furthermore, annotators are required to highlight and identify specific problematic sections within responses to pinpoint exact issues beyond preference annotation.

\begin{table*}[t]
\centering
\begin{tabular}{>{\centering}p{0.1\linewidth}cp{0.15\linewidth}p{0.55\linewidth}}
\toprule
\textbf{Quality} & \textbf{Score} & \textbf{Category} & \textbf{Detailed Description} \\
\midrule
\multirow{2}{*}{\centering Poor} & \centering 0 & \centering Severe Errors & Response contains severe mistakes with no practical value. Examples: harmful content, completely ignored instructions, text collapse, wrong language (non-Chinese), severe content missing (truncated), blank, or error messages. \\
\cmidrule{2-4}
& \centering 1 & \centering Extremely Low Quality & Response performs extremely poorly in all 3H dimensions, with major/numerous errors in format, information, and text. Cannot meet user needs; overall impression is extremely poor and generally unusable. \\
\midrule
\multirow{2}{*}{\centering Average} & \centering 2 & \centering Below Average & Response shows deficiencies in 3H dimensions with some obvious but non-fatal issues. Poor overall impression, most content unusable, small portions might be usable after adjustment. \\
\cmidrule{2-4}
& \centering 3 & \centering Moderate & Response shows average performance in 3H dimensions, basically meets user needs. Contains minor errors with limited impact. Acceptable impression, mediocre, partially adoptable but requires user adjustments. \\
\midrule
\multirow{2}{*}{\centering Excellent} & \centering 4 & \centering Good & Response performs well in 3H dimensions, meets user needs with no hard errors. Good overall impression with minimal flaws or none (but no highlights). Mostly directly usable, small portions need minor adjustments. \\
\cmidrule{2-4}
& \centering 5 & \centering Outstanding & Response excels in all 3H dimensions, exceptional overall impression with brilliant points/highlights. Perfectly addresses user needs; highly suitable for the scenario, entirely adoptable without changes. \\
\bottomrule
\end{tabular}
\caption{Scoring criteria and detailed descriptions for response quality assessment.}
\label{tab:score_criteria}
\end{table*}

\subsection{Annotation Bias}
\label{sec:bias}
\begin{figure*}[ht!]
    \centering
    \begin{subfigure}{0.58\linewidth}
        \centering
        \includegraphics[width=\linewidth]{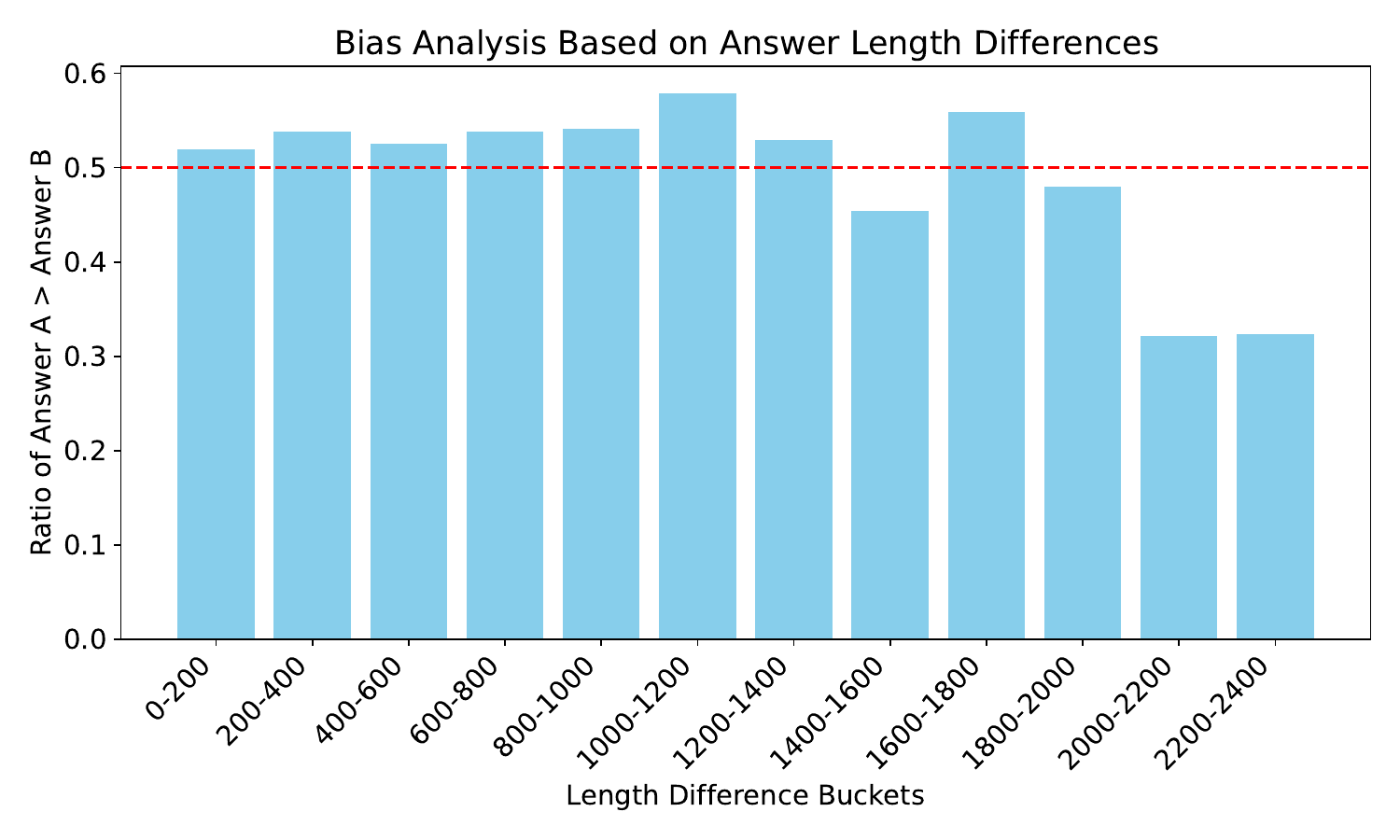}
        \caption{Human Annotator - Length Bias.}
        \label{fig:human_length}
    \end{subfigure}%
    \hfill
    \begin{subfigure}{0.36\linewidth}
        \centering
        \includegraphics[width=\linewidth]{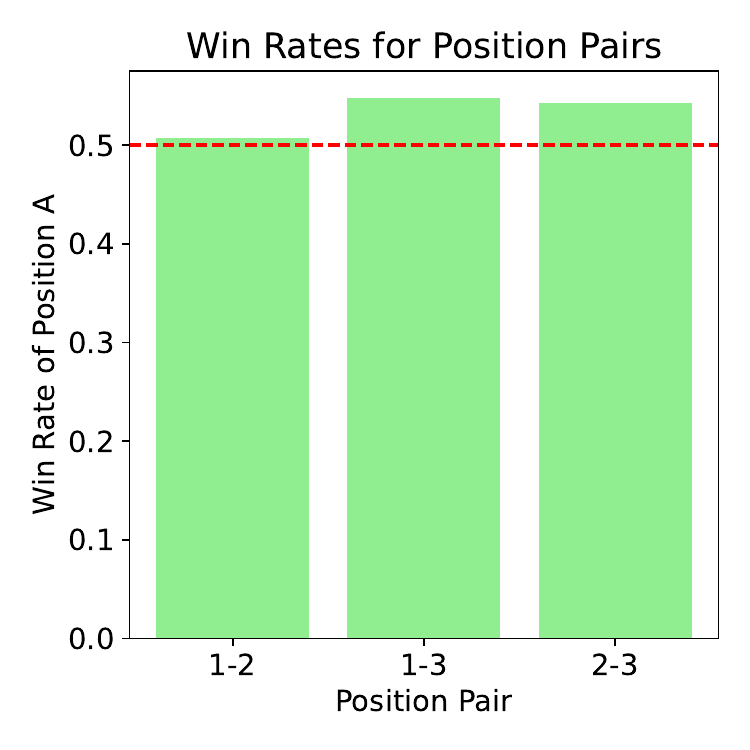}
        \caption{Human Annotator - Positional Bias.}
        \label{fig:human_position}
    \end{subfigure}
    \begin{subfigure}{0.58\linewidth}
        \centering
        \includegraphics[width=\linewidth]{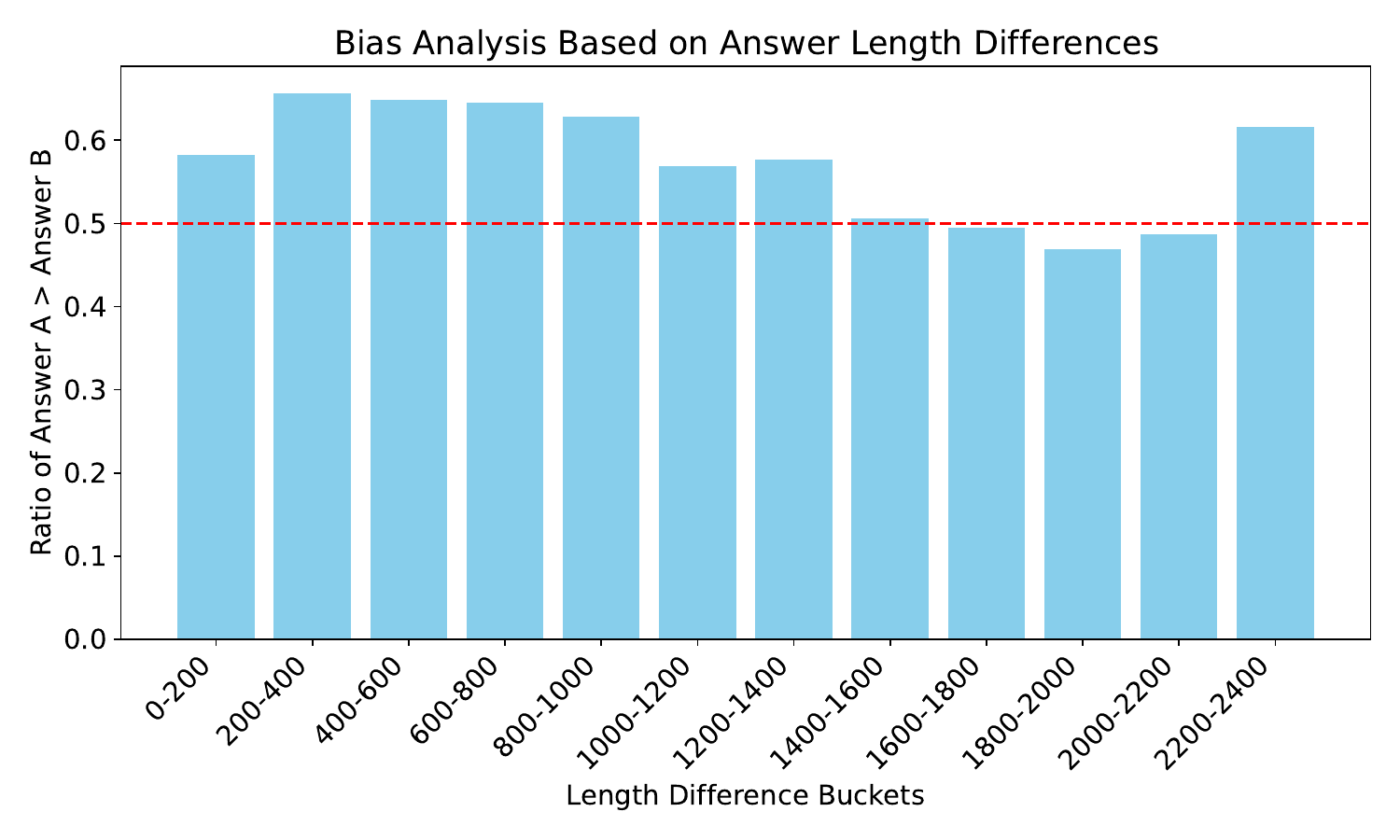}
        \caption{GPT Annotator - Length Bias.}
        \label{fig:gpt_length}
    \end{subfigure}%
    \hfill
    \begin{subfigure}{0.36\linewidth}
        \centering
        \includegraphics[width=\linewidth]{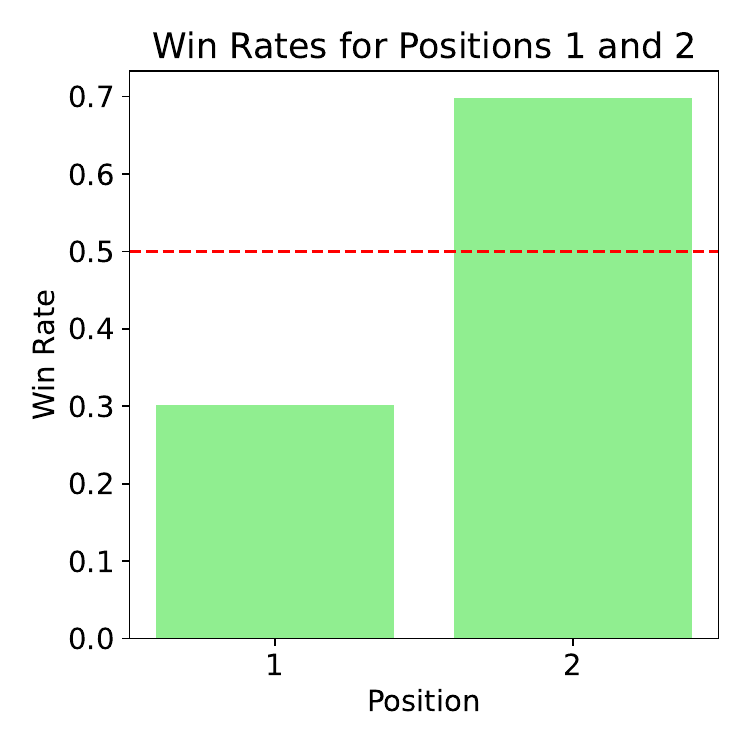}
        \caption{GPT Annotator - Positional Bias.}
        \label{fig:gpt_position}
    \end{subfigure}
    \caption{Comparison of Human and GPT Annotator Biases. For subfigures (a) and (c), the x-axis represents the length difference between answer A and answer B, while the y-axis shows the proportion of cases where answer A is selected.}
    \label{fig:combined_biases}
\end{figure*}

We explore the preferences of both human and GPT annotators in terms of response length and position, as shown in Figure \ref{fig:combined_biases}. 
It can be observed that GPT-4o generally prefers responses that are placed later, whereas human annotators do not exhibit a significant preference for position. Additionally, when the response length difference is moderate, both human and GPT annotators tend to favor longer responses. 
However, as the length difference becomes too large, humans tend to prefer shorter ones. 
Overall, the specific preferences of the annotators are not very pronounced.

\section{Benchmark Results}
\label{sec:benchmark}

\renewcommand{\arraystretch}{0.75}
\begin{table*}[t]
\centering
\setlength{\extrarowheight}{0pt}
\addtolength{\extrarowheight}{\aboverulesep}
\addtolength{\extrarowheight}{\belowrulesep}
\setlength{\aboverulesep}{0pt}
\setlength{\belowrulesep}{0pt}
\resizebox{\linewidth}{!}{%
\begin{tabular}{lcccccc} 
\toprule
\multicolumn{1}{c}{\multirow{2}{*}{\textbf{Model Name}}} & \multicolumn{1}{l}{\multirow{2}{*}{\textbf{RewardBench}}} & \multicolumn{2}{c}{\textbf{Open Prompt}} & \multicolumn{2}{c}{\textbf{Human Instruction}} & \multirow{2}{*}{\textbf{Overall}} \\ 
\cline{3-6}
\multicolumn{1}{c}{} & \multicolumn{1}{l}{} & Acc. & Exact. & Acc. & Exact. & \\ 
\midrule
\multicolumn{7}{c}{\textbf{\textit{Open-source Reward Models}}} \\ 
\midrule
\href{https://huggingface.co/Skywork/Skywork-Reward-Gemma-2-27B}{Skywork-Reward-Gemma-2-27B} & \redcellfull{0.938} & \orangecellfull{0.754} & \yellowcellfull{0.329} & \greencellfull{0.748} & \greenbluecellfull{0.311} & \bluecellfull{0.535} \\
\href{https://huggingface.co/Skywork/Skywork-Reward-Gemma-2-27B-v0.2}{Skywork-Reward-Gemma-2-27B-v0.2} & \redcellfull{0.943} & \orangecellfull{0.751} & \yellowcellfull{0.321} & \greencellfull{0.735} & \greenbluecellfull{0.294} & \bluecellfull{0.525} \\
\href{https://huggingface.co/nvidia/Llama-3.1-Nemotron-70B-Reward-HF}{Llama-3.1-Nemotron-70B-Reward-HF} & \redcellfull{0.941} & \orangecellfull{0.750} & \yellowcellfull{0.317} & \greencellfull{0.722} & \greenbluecellfull{0.271} & \bluecellfull{0.515} \\
\href{https://huggingface.co/NCSOFT/Llama-3-OffsetBias-RM-8B}{Llama-3-OffsetBias-RM-8B} & \redcellfull{0.894} & \orangecellfull{0.734} & \yellowcellfull{0.310} & \greencellfull{0.689} & \greenbluecellfull{0.239} & \bluecellfull{0.493} \\
\href{https://huggingface.co/weqweasdas/RM-Mistral-7B}{RM-Mistral-7B} & \redcellfull{0.804} & \orangecellfull{0.721} & \yellowcellfull{0.285} & \greencellfull{0.700} & \greenbluecellfull{0.259} & \bluecellfull{0.491} \\
\href{https://huggingface.co/LxzGordon/URM-LLaMa-3-8B}{URM-LLaMa-3-8B} & \redcellfull{0.899} & \orangecellfull{0.727} & \yellowcellfull{0.310} & \greencellfull{0.688} & \greenbluecellfull{0.230} & \bluecellfull{0.489} \\
\href{https://huggingface.co/RLHFlow/ArmoRM-Llama3-8B-v0.1}{ArmoRM-Llama3-8B-v0.1} & \redcellfull{0.904} & \orangecellfull{0.715} & \yellowcellfull{0.308} & \greencellfull{0.677} & \greenbluecellfull{0.246} & \bluecellfull{0.487} \\
\href{https://huggingface.co/Skywork/Skywork-Reward-Llama-3.1-8B-v0.2}{Skywork-Reward-Llama-3.1-8B-v0.2} & \redcellfull{0.931} & \orangecellfull{0.721} & \yellowcellfull{0.283} & \greencellfull{0.701} & \greenbluecellfull{0.237} & \bluecellfull{0.486} \\
\href{https://huggingface.co/LxzGordon/URM-LLaMa-3.1-8B}{URM-LLaMa-3.1-8B} & \redcellfull{0.929} & \orangecellfull{0.722} & \yellowcellfull{0.292} & \greencellfull{0.696} & \greenbluecellfull{0.230} & \bluecellfull{0.485} \\
\href{https://huggingface.co/Ray2333/GRM-Llama3-8B-rewardmodel-ft}{GRM-Llama3-8B-rewardmodel-ft} & \redcellfull{0.915} & \orangecellfull{0.728} & \yellowcellfull{0.281} & \greencellfull{0.688} & \greenbluecellfull{0.229} & \bluecellfull{0.482} \\
\href{https://huggingface.co/nicolinho/QRM-Llama3.1-8B}{QRM-Llama3.1-8B} & \redcellfull{0.931} & \orangecellfull{0.722} & \yellowcellfull{0.275} & \greencellfull{0.691} & \greenbluecellfull{0.233} & \bluecellfull{0.480} \\
\href{https://huggingface.co/Skywork/Skywork-Reward-Llama-3.1-8B}{Skywork-Reward-Llama-3.1-8B} & \redcellfull{0.931} & \orangecellfull{0.721} & \yellowcellfull{0.273} & \greencellfull{0.690} & \greenbluecellfull{0.230} & \bluecellfull{0.479} \\
\href{https://huggingface.co/sfairXC/FsfairX-LLaMA3-RM-v0.1}{FsfairX-LLaMA3-RM-v0.1} & \redcellfull{0.844} & \orangecellfull{0.710} & \yellowcellfull{0.286} & \greencellfull{0.667} & \greenbluecellfull{0.224} & \bluecellfull{0.472} \\
\href{https://huggingface.co/weqweasdas/RM-Gemma-7B}{RM-Gemma-7B} & \redcellfull{0.695} & \orangecellfull{0.700} & \yellowcellfull{0.273} & \greencellfull{0.678} & \greenbluecellfull{0.235} & \bluecellfull{0.471} \\
\href{https://huggingface.co/internlm/internlm2-20b-reward}{internlm2-20b-reward} & \redcellfull{0.902} & \orangecellfull{0.714} & \yellowcellfull{0.260} & \greencellfull{0.652} & \greenbluecellfull{0.200} & \bluecellfull{0.457} \\
\href{https://huggingface.co/internlm/internlm2-7b-reward}{internlm2-7b-reward} & \redcellfull{0.876} & \orangecellfull{0.712} & \yellowcellfull{0.262} & \greencellfull{0.644} & \greenbluecellfull{0.187} & \bluecellfull{0.451} \\
\href{https://huggingface.co/CIR-AMS/BTRM_Qwen2_7b_0613}{BTRM-Qwen2-7b-0613} & \redcellfull{0.832} & \orangecellfull{0.708} & \yellowcellfull{0.259} & \greencellfull{0.647} & \greenbluecellfull{0.186} & \bluecellfull{0.450} \\
\href{https://huggingface.co/weqweasdas/RM-Gemma-2B}{RM-Gemma-2B} & \redcellfull{0.654} & \orangecellfull{0.662} & \yellowcellfull{0.222} & \greencellfull{0.633} & \greenbluecellfull{0.205} & \bluecellfull{0.431} \\
\href{https://huggingface.co/internlm/internlm2-1_8b-reward}{internlm2-1-8b-reward} & \redcellfull{0.822} & \orangecellfull{0.642} & \yellowcellfull{0.182} & \greencellfull{0.619} & \greenbluecellfull{0.163} & \bluecellfull{0.402} \\
\href{https://huggingface.co/Ray2333/GRM-llama3-8B-distill}{GRM-llama3-8B-distill} & \redcellfull{0.862} & \orangecellfull{0.531} & \yellowcellfull{0.123} & \greencellfull{0.548} & \greenbluecellfull{0.127} & \bluecellfull{0.332} \\
\href{https://huggingface.co/Ray2333/GRM-Gemma-2B-rewardmodel-ft}{GRM-Gemma-2B-rewardmodel-ft} & \redcellfull{0.845} & \orangecellfull{0.509} & \yellowcellfull{0.111} & \greencellfull{0.470} & \greenbluecellfull{0.106} & \bluecellfull{0.299} \\
\href{https://huggingface.co/Ray2333/Gemma-2B-rewardmodel-ft}{Gemma-2B-rewardmodel-ft} & \redcellfull{0.805} & \orangecellfull{0.494} & \yellowcellfull{0.106} & \greencellfull{0.473} & \greenbluecellfull{0.111} & \bluecellfull{0.296} \\
\href{https://huggingface.co/GRM-gemma2-2B-rewardmodel-ft}{GRM-gemma2-2B-rewardmodel-ft} & \redcellfull{0.884} & \orangecellfull{0.471} & \yellowcellfull{0.093} & \greencellfull{0.480} & \greenbluecellfull{0.110} & \bluecellfull{0.288} \\
\midrule
\multicolumn{7}{c}{\textbf{\textit{Generative Models as Reward Models}}} \\ 
\midrule
\href{https://huggingface.co/Skywork/Skywork-Critic-Llama-3.1-70B}{Skywork-Critic-Llama-3.1-70B} & \redcellfull{0.933} & \orangecellfull{0.755} & \yellowcellfull{0.320} & \greencellfull{0.731} & \greenbluecellfull{0.258} & \bluecellfull{0.516} \\
\href{https://huggingface.co/opencompass/CompassJudger-1-14B-Instruct}{CompassJudger-1-14B-Instruct} & \redcellfull{0.841} & \orangecellfull{0.745} & \yellowcellfull{0.327} & \greencellfull{0.692} & \greenbluecellfull{0.239} & \bluecellfull{0.501} \\
\href{https://huggingface.co/opencompass/CompassJudger-1-32B-Instruct}{CompassJudger-1-32B-Instruct} & \redcellfull{0.852} & \orangecellfull{0.742} & \yellowcellfull{0.322} & \greencellfull{0.685} & \greenbluecellfull{0.231} & \bluecellfull{0.495} \\
\href{https://huggingface.co/Qwen/Qwen2.5-72B-Instruct}{Qwen2.5-72B-Instruct} & - & \orangecellfull{0.734} & \yellowcellfull{0.306} & \greencellfull{0.678} & \greenbluecellfull{0.229} & \bluecellfull{0.487} \\
\href{https://huggingface.co/Skywork/Skywork-Critic-Llama-3.1-8B}{Skywork-Critic-Llama-3.1-8B} & \redcellfull{0.890} & \orangecellfull{0.726} & \yellowcellfull{0.288} & \greencellfull{0.696} & \greenbluecellfull{0.229} & \bluecellfull{0.485} \\
\href{https://openai.com/gpt-4}{GPT4o} & \redcellfull{0.846} & \orangecellfull{0.727} & \yellowcellfull{0.300} & \greencellfull{0.667} & \greenbluecellfull{0.203} & \bluecellfull{0.457} \\
\href{https://agicto.com/model/doubao-pro-128k}{Doubao-pro-128k} & - & \orangecellfull{0.720} & \yellowcellfull{0.280} & \greencellfull{0.662} & \greenbluecellfull{0.164} & \bluecellfull{0.456} \\
\href{https://huggingface.co/Qwen/Qwen2.5-7B-Instruct}{Qwen2.5-7B-Instruct} & - & \orangecellfull{0.713} & \yellowcellfull{0.262} & \greencellfull{0.637} & \greenbluecellfull{0.163} & \bluecellfull{0.444} \\
\href{https://huggingface.co/NCSOFT/Llama-3-OffsetBias-8B}{Llama-3-OffsetBias-8B} & \redcellfull{0.840} & \orangecellfull{0.690} & \yellowcellfull{0.243} & \greencellfull{0.658} & \greenbluecellfull{0.180} & \bluecellfull{0.443} \\
\href{https://huggingface.co/meta-llama/Llama-3.1-70B-Instruct}{Llama-3.1-70B-Instruct} & \redcellfull{0.840} & \orangecellfull{0.685} & \yellowcellfull{0.244} & \greencellfull{0.610} & \greenbluecellfull{0.153} & \bluecellfull{0.423} \\
\href{https://huggingface.co/opencompass/CompassJudger-1-1.5B-Instruct}{CompassJudger-1-1.5B-Instruct} & \redcellfull{0.734} & \orangecellfull{0.660} & \yellowcellfull{0.210} & \greencellfull{0.594} & \greenbluecellfull{0.132} & \bluecellfull{0.399} \\
\href{https://huggingface.co/meta-llama/Llama-3.1-8B-Instruct}{Llama-3.1-8B-Instruct} & \redcellfull{0.657} & \orangecellfull{0.630} & \yellowcellfull{0.158} & \greencellfull{0.583} & \greenbluecellfull{0.116} & \bluecellfull{0.372} \\
\href{https://openai.com/gpt-3.5-turbo}{GPT3.5-turbo} & \redcellfull{0.653} & \orangecellfull{0.616} & \yellowcellfull{0.143} & \greencellfull{0.572} & \greenbluecellfull{0.113} & \bluecellfull{0.361} \\
\bottomrule
\end{tabular}
}
\caption{Performance of discriminative and generative RMs on CheemsBench. The \textbf{Overall} metric is the average of accuracy (\textbf{Acc.}) and exact match (\textbf{Exact.}) across the Open Prompt and Human Instruction subsets.}
\label{tab:rm_results_full}
\end{table*}
\renewcommand{\arraystretch}{1}

In this section, we present comprehensive results on CheemsBench. 
Table \ref{tab:rm_results_full} reports the performance of both discriminative RMs and generative models serving as RMs.
The evaluated discriminative RMs include \textbf{Skywork-series} \cite{liu2024skyworkrewardbagtricksreward}, \textbf{Llama-3.1-Nemotron-70B-Reward} \cite{wang2024helpsteer2preferencecomplementingratingspreferences}, \textbf{Llama-3-OffsetBias-RM-8B} \citep{park2024offsetbias}, \textbf{RM-Mistral-7B} \citep{xiong2024iterativepreferencelearninghuman}, \textbf{URM-series} \citep{lou2024uncertainty},  \textbf{ArmoRM-Llama3-8B-v0.1} \citep{ArmoRM}, \textbf{GRM-series} \citep{yang2024regularizing}, \textbf{QRM-series} \citep{dorka2024quantile}, \textbf{FsfairX-LLaMA3-RM-v0.1} \citep{dong2023raft}, \textbf{RM-Gemma-2/7B} \citep{dong2023raft}, \textbf{IntermLM-series} \citep{cai2024internlm2technicalreport}, \textbf{BTRM-Qwen2-7b-0613}.
The evaluated generative models as RMs include \textbf{Skywork-Critic-series} \cite{liu2024skyworkrewardbagtricksreward}, \textbf{CompassJudger-Series} \citep{cao2024compass}, \textbf{Qwen2.5-Series} \citep{qwen2.5}, \textbf{Llama3.1-Series} \citep{grattafiori2024llama3herdmodels}, \textbf{Llama-3-OffsetBias-8B} \citep{park2024offsetbias}.
For commercial models like \textbf{GPT-4}, \textbf{GPT-3.5-turbo} and \textbf{Doubao-pro}, we use their official APIs for evaluation.
Table \ref{tab:preference_data} reports the performance of different datasets.
The evaluated datasets include \textbf{HH-RLHF-cn}, \textbf{Huozi} \citep{huozi}, \textbf{Kyara} \citep{Yang_Kyara_2024}, \textbf{Zhihu}, \textbf{ChatbotArene} \citep{zheng2023judging}, HH-RLHF \citep{ganguli2022redteaminglanguagemodels}, \textbf{MathPreference}, \textbf{Nectar} \citep{starling2023}, \textbf{PKU-SafeRLHF} \citep{ji2024pku}, Skywork \citep{liu2024skyworkrewardbagtricksreward}, MathStackExchange \citep{h4stackexchange}, \textbf{UltraFeedback} \citep{cui2023ultrafeedback}, \textbf{HelpSteer2} \citep{wang2024helpsteer2}.


\section{Hyperparameter Settings}
\label{sec:hyper}
We present the key hyperparameters used in our experiments in Table \ref{tab:hyperparameters}. Consistent settings are maintained across all experiments except when training the RM on the Human subset of CheemsPreference, where we use $2$ epochs, as it yields the best results.
We report the experiment results for a single run.

\begin{table}[H]
\centering
\resizebox{0.9\linewidth}{!}{%
\begin{tabular}{ll} 
\toprule
\textbf{Hyperparameter} & \textbf{Value} \\ 
\hline
Max Sequence Length & 2048 \\
Regularization Coefficient & 0.1 \\
Gradient Accumulation Steps & 4 \\
Micro Batch Size & 2 \\
Global Batch Size & 256 \\
Epochs & 2 \\
Warmup Ratio & 0.1 \\
Learning Rate Scheduler & Cosine \\
Learning Rate & 5e-6 \\
\bottomrule
\end{tabular}
}
\caption{Hyperparameter settings.}
\label{tab:hyperparameters}
\end{table}

\section{Use of AI Assistants}
We use AI to assist with grammar checks, sentence polish and coding.


\end{document}